\pdfoutput = 1
\documentclass[twocolumn]{article}

\usepackage{times}
\usepackage[numbers]{natbib}
\usepackage[english]{babel}
\usepackage{blindtext}
\usepackage{graphicx}
\usepackage{amsmath, amsthm, amssymb, bbm, bm}
\usepackage{enumerate}
\usepackage{float}      
\usepackage{subcaption}  
\usepackage{wrapfig}
\usepackage[margin=0cm]{caption}
\usepackage[titletoc, toc]{appendix}
\usepackage{tabularx}

\usepackage{multirow}
\usepackage{hhline}
\usepackage{makecell}
\usepackage{placeins}  

\usepackage[x11names, usenames, dvipsnames, svgnames, table]{xcolor}
\definecolor{firebrick}{rgb}{.698,.133,.133}
\definecolor{mybluelight}{rgb}{0.9, 0.9, 1.}
\definecolor{myorangelight}{rgb}{1., 0.9, 0.9}

\usepackage[utf8]{inputenc} 
\usepackage[T1]{fontenc}    
\usepackage{url}            
\usepackage{booktabs, colortbl}       
\usepackage{amsfonts}       
\usepackage{nicefrac}       
\usepackage{microtype}      

\usepackage{csquotes}
\usepackage{latexsym}

\usepackage{pifont}
\newcommand{\cmark}{\ding{51}}%
\newcommand{\xmark}{\ding{55}}%

\usepackage[boxruled, vlined, linesnumbered]{algorithm2e}
\SetAlFnt{\small}
\SetAlCapFnt{\small}
\SetAlCapNameFnt{\small}
\usepackage{algorithmic}
\algsetup{linenosize=\tiny}

\let\oldnl\nl
\newcommand{\nonl}{\renewcommand{\nl}{\let\nl\oldnl}}

\usepackage{paralist}

\usepackage{xspace}
\usepackage{soul}
\usepackage{dsfont}
\usepackage{stmaryrd}
\usepackage[textwidth=15mm]{todonotes}
\usepackage{dirtytalk}
\usepackage{pbox}
\usepackage{cprotect}

\usepackage{verbatim}
\usepackage{textcomp}
\usepackage[normalem]{ulem}

\usepackage{mathtools}
\usepackage{etextools}
\usepackage[inline]{enumitem}

\usepackage[colorlinks=true,allcolors=firebrick,bookmarks=false]{hyperref}

\definecolor{darkergreen}{RGB}{21, 152, 56}
\definecolor{red2}{RGB}{252, 54, 65}
\definecolor{Gray}{gray}{0.85}
\newcolumntype{g}{>{\columncolor{Gray}}c}

\let\OLDthebibliography\thebibliography
\renewcommand\thebibliography[1]{
  \OLDthebibliography{#1}
  \setlength{\parskip}{0pt}
  \setlength{\itemsep}{0pt plus 0.3ex}
}

\newcommand{\reals}{\mathbb{R}}

\newcommand{\abs}[1]{\ensuremath \left| #1 \right|}

\theoremstyle{definition}

\DeclarePairedDelimiterX{\divx}[2]{(}{)}{%
  #1\;\delimsize\|\;#2%
}

\newcommand*{\eg}{\emph{e.g.}\@\xspace}
\newcommand*{\ie}{\emph{i.e.}\@\xspace}

\usepackage{style}

\usepackage[boxruled, vlined, linesnumbered]{algorithm2e}
\SetAlFnt{\small}
\SetAlCapFnt{\small}
\SetAlCapNameFnt{\small}
\usepackage{algorithmic}
\algsetup{linenosize=\tiny}

\newcommand\corloc{\texttt{CorLoc}\xspace}

\newcommand\ytovone{\texttt{YTOv1}\xspace}
\newcommand\ytovtwodtwo{\texttt{YTOv2.2}\xspace}
\newcommand\colocam{\texttt{CoLo-CAM}\xspace}

\newcommand\labelgt{\texttt{GT}\xspace}  
\newcommand\labelpr{\texttt{PR}\xspace}  
\newcommand\labelavg{\texttt{AVG}\xspace}  

\makeatletter
\@namedef{ver@everyshi.sty}{}
\newcommand{\removelatexerror}{\let\@latex@error\@gobble}

\title{\colocam: Class Activation Mapping for Object Co-Localization in \\ Weakly-Labeled Unconstrained Videos}

\renewcommand\footnotemark{}

\author{Soufiane~Belharbi$^{1}$,
  ~Shakeeb~Murtaza$^{1}$,
  ~Marco~Pedersoli$^{1}$,
  ~Ismail~Ben~Ayed$^{1}$,
  ~Luke~McCaffrey$^{2}$, and
  ~Eric~Granger$^{1}$\\
 	$^1$ LIVIA, ILLS, Dept. of Systems Engineering, ÉTS, Montreal, Canada \\
	$^2$ Goodman Cancer Research Centre, Dept. of Oncology, McGill University, Montreal, Canada\\
{\tt\footnotesize \textcolor{black}{\{soufiane.belharbi,Marco.Pedersoli,Ismail.BenAyed,Eric.Granger\}@etsmtl.ca}}\\
{\tt\footnotesize \textcolor{black}{shakeeb.murtaza.1@ens.etsmtl.ca, luke.mccaffrey@mcgill.ca}}
}

\newcommand{\ignore}[1]{}

\begin{document}
\maketitle\thispagestyle{fancy}

\maketitle
\rhead{\color{gray} \small Belharbi et al. \;  [Pattern Recognition 2025, \href{https://doi.org/10.1016/j.patcog.2025.111358}{10.1016/j.patcog.2025.111358}]}

\begin{abstract}
Leveraging spatiotemporal information in videos is critical for weakly supervised video object localization (WSVOL) tasks. However, state-of-the-art methods only rely on visual and motion cues, while discarding discriminative information, making them susceptible to inaccurate localizations. Recently, discriminative models have been explored for WSVOL tasks using a temporal class activation mapping (CAM) method. Although their results are promising, objects are assumed to have limited movement from frame to frame, leading to degradation in performance for relatively long-term dependencies. 
This paper proposes a novel CAM method for WSVOL that exploits spatiotemporal information in activation maps during training without constraining an object's position. Its training relies on \emph{Co}-\emph{Lo}calization, hence, the name \colocam.
Given a sequence of frames, localization is jointly learned based on color cues extracted across the corresponding maps, by assuming that an object has similar color in consecutive frames. CAM activations are constrained to respond similarly over pixels with similar colors, achieving co-localization. This improves localization performance because the joint learning creates direct communication among pixels across all image locations and over all frames, allowing for transfer, aggregation, and correction of localizations. Co-localization is integrated into training by minimizing the color term of a conditional random field (CRF) loss over a sequence of frames/CAMs. 
Extensive experiments\footnote{Code: \href{https://github.com/sbelharbi/colo-cam}{https://github.com/sbelharbi/colo-cam}.} on two challenging YouTube-Objects datasets of unconstrained videos show the merits of our \colocam method, and its robustness to long-term dependencies, leading to new state-of-the-art performance for WSVOL task.
\end{abstract}


\begin{figure*}[ht!]
\centering
  \centering
  \includegraphics[width=.6\linewidth]{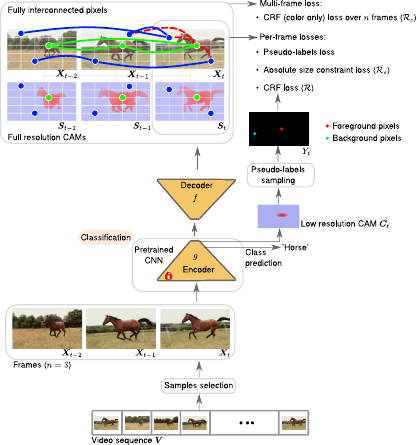}
  \caption[Caption]{Illustration of the combination of the single and multi-frame training process using our \colocam method in the case where $n=3$ frames. For the multi-frame loss, each pair of pixels across all three images are interconnected. For clarity, only few pixels and connections are shown. \textcolor{green}{Green dots} are foreground pixels. \textcolor{blue}{Blue dots} are background pixels. 
  The classifier (encoder $g$ + classification head) is pre-trained, and frozen, and only the decoder $f$ is trained. We employ per-frame and multi-framer losses. Our multi-frame loss ${\mathcal{R}_c}$ assumes that an object of interest has similar colour over multiple adjacent frames, i.e. the object 'horse' in the frames. This assumption also applies to background objects, i.e., the 'sky', 'grass', and 'trees'. CAMs are constrained to have similar responses over pixels with similar color across different locations over all frames. Each pair of pixels across the three frames are interconnected. Solid connections illustrate strong visual similarity, while dotted connections indicate weak similarity. Additionally, the per-frame terms consist of the pseudo-label loss, absolute size constraint (${\mathcal{R}_s}$), and CRF loss (${\mathcal{R}}$).
  }
  \label{fig:proposal}
\end{figure*}

\section{Introduction} \label{sec:intro}

Techniques for automated video analysis have become important due to the recent progress of online multimedia platforms such as YouTube that provide easy access to a large number of videos.  
In this work, we focus on the task of localizing objects in videos, which plays a critical role in understanding video content and improving downstream tasks such as action recognition~\cite{Zhang24}, video-based summarization~\cite{ZhuZHW23}, 
and event detection~\cite{Shao23}, 
object detection~\cite{ShaoCSJXYZX22}, 
facial emotion recognition~\cite{xue22,Zhang22}, 
and visual object tracking~\cite{LuoXMZLK21}. 
Unfortunately, the video annotations required for fully supervised localization, \ie, bounding boxes for each frame, are costly due to the large number of frames. As an alternative, weak supervision using a video class label or tag associated with the entire video is currently the most common annotation for weakly supervised video object localization (WSVOL)~\cite{JerripothulaCY16,tsai2016semantic}\footnote{The WSVOL task is not to be confused with Weakly-supervised Temporal Action Localization (WTAL) task~\cite{ren23} as they both aim to localize an object-class in videos using only video-class label. However, WTAL task aims to predict time segment of an action class while WSVOL aims for spatial localization (bounding boxes) of objects within a frame in a video.}.  
A video class label describes the main object appearing in the video without spatiotemporal information, \ie, spatial location, and start/end time of the object's appearance, which adds more uncertainty to the label.  
Additionally, unconstrained videos are often captured in the wild, with varying quality, moving objects and cameras, changing viewpoints and illumination, and decoding artifacts. This makes  localization from weakly-labeled unconstrained videos a challenging task.

State-of-the-art WSVOL methods often follow a similar strategy. Localization is often formulated as an optimization problem solved over the entire video. Initial segments/proposals are generated based on visual or motion cues, which are then used to identify and refine object locations through post-processing~\cite{Kwak2015,prest2012learning,Tokmakov2016,zhang2020spftn}. This process is optimized under visual appearance and motion constraints to ensure consistency. Other methods rely on co-localization/co-segmentation across videos/images via graphs that allow for a larger view and interactions between segments~\cite{ChenCC12,JerripothulaCY16,joulin2014,tsai2016semantic}. Despite their success, these methods have several limitations. For instance, initial proposals are not necessarily discriminative, \ie, aligned with the video class, due to unsupervised region extraction using visual and motion cues (e.g., optical flow). Moreover, video class labels are typically only used to cluster videos, which further limits the beneficial impact of discriminative information. Additionally, a per-video model is often optimized, leading to several issues, including scalability, inference time, and deployment. Finally, multi-stage training combines the best local solutions, which does not necessarily lead to the best global solution.

Recently, the temporal class activation mapping (TCAM) method has been proposed to train discriminative multi-class deep learning (DL) model for WSVOL~\cite{tcamsbelharbi2023}. This paper focuses on CAM methods for weakly supervised object localization approach (WSOL) that have been successful on still images~\cite{choe2020evaluating,rony2023deep}. Using only global image class labels, CAM-based methods allow training a discriminative multi-class DL model (e.g., a CNN or vision transformer) to classify an image and localize the corresponding object of interest. Strong CAM activations indicate the potential presence of an object, making them particularly suitable for object localization with low-cost weak annotations~\cite{zhou2016learning}. The TCAM method~\cite{tcamsbelharbi2023} relies on pixel-wise pseudo-labels collected from a pre-trained classifier. Built on the F-CAM architecture~\cite{belharbi2022fcam}, TCAM relied on a temporal max-pooling module over consecutive CAMs to improve pixel-wise pseudo-labels.
Despite its success and fast inference, TCAM assumes that objects have limited displacement over frames~\cite{tcamsbelharbi2023}. This can only be beneficial for short time dependencies over a few frames.

To alleviate this issue while benefiting from CAM methods, the \colocam method is proposed for WSVOL (Figs. \ref{fig:proposal-details} and \ref{fig:proposal}). This new multi-class discriminant approach leverages spatiotemporal information without constraining object movement. In particular, \colocam performs \emph{co-localization} via explicit joint learning of CAMs across multiple frames in a video. Under the assumption that an object maintains a similar color \emph{locally}, \colocam achieves co-localization by constraining a sequence of CAMs to be \emph{consistent} by pushing them to activate similarly over pixels with a similar color. To avoid major changes in color/contrast of object appearance, and therefore maintain the color assumption, our joint learning is applied locally over consecutive frames. 
Unlike TCAM~\cite{tcamsbelharbi2023}, \colocam does not assume a limited displacement of objects. Rather, objects can appear anywhere in the image, making for a more flexible method, and leading to a consistent localization.

Our total training loss is composed of per-frame and multi-frame terms, as illustrated in Fig.\ref{fig:proposal-details}, which are optimized simultaneously via standard Stochastic Gradient Descent (SGD). Our multi-frame optimization method is achieved by constructing a fully connected graph between all pairs of pixels across a set of frames. This graph accounts for pixel-wise information in terms of image color and discriminative CAM response. In particular, pixel probability of CAMs is used as unary potentials, while the \emph{color} difference between every two pixels is considered as pairwise potentials. We then minimize a conditional random field (CRF) loss~\cite{tang2018regularized} using only color similarity,  and without spatial information. This allows for co-localization of an object of interest across multiple frames and independently of its location. 
Following TCAM~\cite{tcamsbelharbi2023}, \colocam also relies on standard per-frame local constraints aiming to stimulate and initiate localization. This includes pixel-wise pseudo-labels, and global size constraints, in addition to standard CRF loss based on color and location~\cite{tang2018regularized} in a single image. Since  pixel-wise pseudo labels are not reliably extracted, our co-localization training allows for the transfer, aggregation, and correction of localizations across frames.

\textbf{Our main contributions are summarized as follows:}

\noindent \textbf{(1)} A \colocam method is introduced for WSVOL in unconstrained videos. 
Unlike the state-of-the-art TCAM method~\cite{tcamsbelharbi2023}, our novel temporal training term does not assume that an object maintains a similar location in different frames. Instead, it allows the object to be located anywhere, making it more robust to motion. This is achieved by performing co-localization over a sequence of CAMs constrained to yield similar activation over pixels with a similar color. To that end, \colocam builds a fully connected graph between all pair of pixels to optimize a CRF loss using color.

\noindent \textbf{(2)}  Our extensive experiments and ablations on two challenging public datasets of unconstrained videos, YouTube-Objects~\cite{prest2012learning,KalogeitonFS16}, show the merits of our proposed method and its robustness to long dependency. Most importantly, our produced CAMs become more discriminative. This translates into several advantageous properties including sharp, more complete, and less noisy activations localized more precisely around objects. Moreover, small and large objects are more accurately localized. \colocam provides a new state-of-the-art localization performance on YouTube-Objects datasets. Additionally, our results confirm the potential of discriminative learning for WSVOL tasks, in contrast to relying only on motion and visual cues.

\section{Related Work}
\label{sec:related-w}
 
This section reviews weakly supervised methods for localization, segmentation, and co-localization/co-segmentation in videos trained using only video class as supervision. Additionally, CAM methods for WSOL on images are discussed, as they relate to WSVOL.

\noindent \textbf{Localization.}
Different methods exploit potential proposals as pseudo-supervision for localization~\citep{prest2012learning,zhang2020spftn}. In~\citep{prest2012learning}, class-specific detectors are proposed. First, segments of coherent motion~\citep{BroxM10} are extracted, then a spatiotemporal bounding box is fitted to each segment forming tubes. A single tube is jointly selected by minimizing energy based on the similarity between tubes, visual homogeneity over time, and the likelihood to contain an object. The selected tubes are used as box supervision to train a per-class detector~\citep{FelzenszwalbGMR10}. In~\citep{zhang2020spftn}, a DL model is trained using noisy supervision to perform segmentation and localization. Given a set of videos from the same class, bounding boxes and segmentation proposals are estimated using advanced optical flow~\citep{LeeKG11}. 
Other methods seek to discover a prominent object (foreground) in (a) video(s)~\citep{jun2016pod,RochanRBW16,Kwak2015}. Then, similar regions are localized and refined using visual appearance and motion consistency. In~\citep{RochanRBW16}, box proposals are generated in a video, and only the relevant ones are retained for building an object appearance model. Maximum a posteriori inference over an undirected chain graphical model is employed to enforce the temporal appearance consistency between adjacent frames in the same video. 
The method in~\citep{Kwak2015} leverages two simultaneous and complementary processes for object localization: the discovery of similar objects in different videos, and the tracking of prominent regions in individual videos. Region proposals~\citep{ManenGG13}, in addition to appearance and motion confidence, are used to determine foreground objects~\citep{BroxM10}, and to maintain temporal consistency between consecutive frames.

\noindent \textbf{Segmentation.} 
%
Independent spatiotemporal segments are first extracted~\citep{HartmannGHTKMVERS12,tang2013discriminative,XuXC12,YanXCC17} via unsupervised methods~\citep{BanicaAIS13,XuXC12} or with a proposals generator~\citep{zhang2015semantic} using pretrained detectors~\citep{zhang2015semantic}. Then, using multiple properties such as visual appearance, and motion cues, these segments are labeled, while preserving temporal consistency. Graphs, such as Conditional Random Field (CRF) and GrabCut approaches~\citep{RotherKB04}, are often employed to ensure consistency. 
In~\citep{LiuTSRCB14} deal with multi-class video segmentation is proposed, where a nearest neighbor-based label transfer between videos is exploited. First, videos are first segmented into spatiotemporal supervoxels~\citep{XuXC12}, and represented by high-dimensional feature vectors using color, texture, patterns, and motion. These features are compressed using binary hashing for semantic similarity measures. A label transfer algorithm is designed using a graph that fosters label smoothness between spatiotemporal adjacent supervoxels in the same video, and between visually similar supervoxels across videos.
In M-CNN~\citep{Tokmakov2016}, a fully convolutional network (FCN) is combined with motion cues to estimate pseudo-labels. Using motion segmentation, a Gaussian mixture model is exploited to estimate foreground appearance potentials~\citep{papazoglou2013fast}. Combined with the FCN prediction, these potentials are used to yield better segmentation pseudo-labels via a GrabCut-like method~\citep{RotherKB04}, and these are then used to fit the FCN.
The previously mentioned segmentation methods use video tags to cluster videos with the same class. While other methods do not rely on such supervision, the process is generally similar. Foreground regions are first estimated~\citep{Croitoru2019,Halle2017,papazoglou2013fast,umer2021efficient} using motion cues~\citep{SundaramBK10}, or PCA, or VideoPCA~\citep{StretcuL15}. This initial segmentation is later refined via graph-methods~\citep{RotherKB04}.

\noindent \textbf{Co-segmentation/Co-localization.}
The co-segmentation methods have been leveraged to segment similar objects across a set of videos. Typical methods rely on discerning common objects via the similarity of different features including visual appearance, motion, and shape. Using initial guessing of objects/segments, graphs, such as CRF and graph cuts, are considered to model relationships between them~\citep{ChenCC12,FuXZL14,tsai2016semantic,ZhangJS14}.  In~\citep{ChenCC12}, the authors use relative intra-video motion extracted from dense optical flow, and inter-video co-features based on Gaussian Mixture Models (GMM). The common object, \ie, the foreground, in two videos is isolated from the background using a Markov Random Field (MRF). It is solved iteratively via graph cuts while accounting for a unary term based on the distance to GMMs, and a pairwise energy based on feature distance weighted by the relative motion distance between super-voxels. The authors in~\citep{FuXZL14} pursue a similar path. The method starts by generating class-independent proposals~\citep{EndresH10}. To identify the foreground object in each frame, various object characteristics are used while accounting for inter- and intra- video coherence of the foreground. A co-selection graph is formulated as a CRF exploiting unary energy from motion (optical flow), and pairwise energy over the foreground that combines visual (histogram) and shape similarities. In~\citep{ZhangJS14}, co-segmentation in videos is performed by sampling, tracking, and matching object proposals using graphs. It prunes away noisy segments in each video by selecting proposal tracklets that are spatially salient and temporally consistent. An iterative grouping process is used to gather objects with similar shapes and appearances (histogram) within and across videos. The final object regions obtained are used to initialize a segmentation process~\citep{FulkersonVS09}. The authors in~\citep{tsai2016semantic} use a pre-trained FCN to generate object-like tracklets which are linked for each object category via a graph. To define the corresponding relation between tracklets, a sub-modular optimization is formulated  based on the similarities of the tracklets via object appearance, shape, and motion. To discover prominent objects in each video, tracklets are ranked based on their mutual similarities.

Other methods aim to co-localize objects in images or videos using bounding boxes instead of segments. In~\citep{joulin2014}, a large number of proposed bounding boxes is generated by relying on objectness~\citep{AlexeDF12} with the goal being to select a single box that is more likely to contain the common object across images/videos. This is achieved by solving a quadratic problem using the Frank-Wolfe algorithm~\citep{Frank1956AnAF}. Following~\citep{TangJLF14}, box similarity and discriminability are used as constraints. Additionally, to ensure consistency between consecutive frames, multiple object properties including visual appearance, position, and size are used. \citep{JerripothulaCY16} leverage co-saliency as a prior to filter out noisy bounding box proposals. The co-saliency map is derived from inter- and intra-video commonness, and total motion saliency maps. Potential proposals are used for tracklet generation, and tracklets with a high confidence score and exhibiting good spatiotemporal consistency yield the final localization.

\noindent \textbf{WSOL in still images.}
CAM-based methods have emerged as a dominant approach for the WSOL~\citep{choe2020evaluating,murtaza25,rony2023deep} on still images. 
Early works focused on designing variant spatial pooling layers~\citep{durand2017wildcat,durand2016weldon, lin2013network, oquab2015object, pinheiro2015image,sun2016pronet,zhou2016learning, ZhouZYQJ18PRM}. An extension to a multi-instance learning framework (MIL) has been considered~\citep{ilse2018attention}. However, CAMs tend to under-activate by highlighting only small and the most discriminative parts of an object~\citep{choe2020evaluating,rony2023deep}, diminishing its localization performance. To overcome this, different strategies have been considered, including data augmentation over input image or deep features~\citep{belharbi2020minmaxuncer,ChoeS19,LiWPE018CVPR,MaiYL20eil,SinghL17,wei2017object,YunHCOYC19,ZhangWF0H18,zhu2017soft}, as well as architectural changes~\citep{gao2021tscam,KiU0B20icl,LeeKLLY19,XueLWJJY19iccvdanet,YangKKK20,ZhangW020i2c}. Recently, learning via pseudo-labels shown some potential, despite its reliance on noisy labels~\citep{belharbi24-fer-aus,belharbi2022fcam,negevsbelharbi2022,MeethalPBG20icprcstn,murtaza24tedloc,murtaza2022dipssypo,murtaza2022dips,wei2021shallowspol,ZhangCW20rethink,ZhangWKYH18}. Most previous methods used only forward information in a CNN, with some models designed to leverage backward information as well. These include biologically inspired methods~\citep{cao2015look,zhang2018top}, and gradient~\citep{ChattopadhyaySH18wacvgradcampp,fu2020axiom,JiangZHCW21layercam,SelvarajuCDVPB17iccvgradcam} or confidence score aggregation methods~\citep{desai2020ablation,naidu2020iscam,naidu2020sscam,WangWDYZDMH20scorecam}. 
Recently, prompt-based methods such as CLIP-ES~\cite{lin2023} trained using text and images allow the prediction of ROIs related to a specific input class. However, such models lack direct classification capability. 
While CAM-based methods are successful on still images, they still require adaptation to leverage the temporal information in videos. Authors in~\cite{tcamsbelharbi2023} considered improving the quality of pseudo-labels to train an F-CAM architecture~\cite{belharbi2022fcam}. To this end, they aggregate CAMs extracted from a sequence of frames into a single CAM that covers more ROIs. However, this led to limited success since degradation in localization performance has been reported after a sequence of 2 frames.

Despite the reported success, state-of-the-art WSVOL methods still suffer different limitations. Proposal generation is a critical step often achieved through unsupervised learning using color cues and motion~\cite{Tokmakov2016}. Additionally, the discriminative information, \ie, video class label, is not used directly, but rather to cluster videos~\cite{JerripothulaCY16,Kwak2015,zhang2020spftn,joulin2014}. As such, the initial localization is vulnerable to errors, which can hinder localization performance in subsequent steps. Moreover, localization is often formulated as the solution to an optimization problem over a single/cluster of videos~\cite{JerripothulaCY16,tsai2016semantic,joulin2014}. This requires a model per video/cluster of videos. Therefore, the operational deployment of WSVOL methods at inference time is expensive and impractical for real-world applications, and does not scale well to large number of classes.

\noindent \textbf{Relation to F-CAM~\cite{belharbi2022fcam} and TCAM~\cite{tcamsbelharbi2023}}: 
F-CAM~\cite{belharbi2022fcam} is a WSOL method for still images. While it does not leverage temporal information, its training relies on key terms to alleviate standard CAM issues such as CAM under-activation and bloby effect~\cite{rony2023deep}. It performs classification and it produces full size CAM, thanks to its U-Net like architecture, allowing the application of different pixel-wise losses, such as pixel pseudo-labels and CRF~\cite{tang2018regularized}.
The recent TCAM method~\cite{tcamsbelharbi2023} builds on top of F-CAM. It uses the same F-CAM architecture, and its same per-frame losses. It aims to leverage temporal information to improve localization. This is done by improving pixel pseudo-labels accuracy via a new temporal max-pooling over low resolution CAMs from a pretrained classifier. It allows to gather relevant spatial ROIs through time. However, its localization performance declines when increasing time dependency, a consequence of the TCAM assumption about the limited movement of objects. This leads to noisy pseudo-labels and, therefore, poor localization. 
Our \colocam method also builds on top of F-CAM architecture, and its per-frame losses. However, the major difference is how to leverage temporal information to improve localization in videos. In this paper, a new constraint loss is proposed for joint training over a sequence of frames to perform co-localization. We explicitly constrain the output CAMs to respond similarly over objects with similar colors in a sequence of frames. This assumes objects have similar colors in adjacent frames. Meanwhile, no position assumption is made allowing objects to move freely which adds more flexibility. The interconnection created between frames/CAMs allows for knowledge transfer and correction of localization. The resulting discriminative CAMs provide a more precise object coverage and less noisy activations.

\section{Proposed Approach}
\label{sec:method}

Let us denote by ${\mathbb{D} = \{(\bm{V}, y)_i\}_{i=1}^N}$ a training set of videos, where ${\bm{V} = \{\bm{X}_t\}_{t=1}^{T}}$ is a video with $T$ frames, ${\bm{X}_t: \Omega \subset \reals^2}$ is the $t$-th frame, and ${\Omega}$ is a discrete image domain. The global video class label is denoted ${y \in \{1, \cdots, K\}}$, with $K$ being the total number of classes. We assume that the label $y$ is assigned to all frames within a video.  Our model follows an F-CAM model~\cite{belharbi2022fcam} which has a U-Net style architecture with skip connections (Fig.\ref{fig:proposal-details}). It is composed of two parts: a standard encoder ${g}$ with parameters ${\bm{\theta^{\prime}}}$ for image classification and a standard decoder ${f}$ with parameters ${\bm{\theta}}$ for localization. 

Our classifier ${g}$ is composed of a backbone encoder for extracting features, as well as a classification scoring head. Classification probabilities per class are denoted ${g(\bm{X}) \in [0, 1]^K}$ where ${g(\bm{X})_k = \mbox{Pr}(k | \bm{X})}$. This module is trained to classify independent frames via standard cross-entropy: ${\min_{\bm{\theta^{\prime}}} \;  - \log(\mbox{Pr}(y | \bm{X}))}$. Once trained, its weights, ${\bm{\theta^{\prime}}}$, are then frozen and used to produce \emph{classifier low resolution CAM} which is then interpolated to image full size. This CAMs is denoted ${\bm{C}_t}$ for the frame ${\bm{X}_t}$. It is extracted using the video true label $y$~\cite{choe2020evaluating}. Therefore,  ${\bm{C}_t}$ is made semantically consistent with the video class. Later, it is used to generate pseudo-labels to train the decoder ${f}$.

The localizer ${f}$ is a decoder that produces two full-resolution CAMs normalized with softmax, and denoted ${\bm{S}_t = f(\bm{X}_t) \in [0, 1]^{\abs{\Omega} \times 2}}$. ${\bm{S}^0_t, \bm{S}^1_t}$ refer to the background and foreground maps, respectively. Pixel-wise training losses will be applied over these two maps ${\bm{S}_t}$, and they will be used for localization at inference time. Note that ${\bm{S}_t}$ CAMs is produced by the decoder, while ${\bm{C}_t}$ is produced by the classifier.
Let ${\bm{S}_t(p) \in [0, 1]^2}$ denotes a row of matrix ${\bm{S}_t}$, with index ${p \in \Omega}$ indicating a point within ${\Omega}$. We denote by ${\bm{F}^{t, n} = \{\bm{X}_{t-n-1}, \cdots, \bm{X}_{t-1}, \bm{X}_t\}}$ a set of $n$ consecutive frames going back from time $t$, and ${\bm{M}^{t, n} = \{\bm{S}_{t-n-1}, \cdots, \bm{S}_{t-1}, \bm{S}_t\}}$ is its corresponding set of CAMs at the decoder output, in the same order. We consider terms at two levels to train our decoder: per-frame and across-frames. The per-frame terms stimulate localization at the frame level, and ensure its local consistency. However, the multi-frame term coordinates localization and ensures its consistency across multiple frames.

\subsection{Per-Frame Losses} 
\label{subsec:perframe-losses}
At frame level, we adopt three common terms from F-CAM~\cite{belharbi2022fcam} work to ensure local localization consistency. The same terms have been used in T-CAM~\cite{tcamsbelharbi2023} work as well. This includes: pixel pseudo-labels, generic size prior, and standard CRF loss~\cite{tang2018regularized}. This has shown to be effective to reduce several limitations of standard CAMs such as bloby effect, unbalanced activations, and misalignment with object boundaries~\cite{tcamsbelharbi2023,choe2020evaluating,rony2023deep,belharbi2022fcam}. Using pixel pseudo-labels allows to stimulate localization at frame level. We review these three terms briefly in the following paragraph.

\noindent \textbf{a) Learning via pseudo-labels (PLs).} It is commonly known that strong activations in a CAM are more likely to be a foreground, while low activations are assumed to be a background~\cite{zhou2016learning}. Following this, pixel-wise PLs for foreground and background regions are generated using the CAM ${\bm{C}_t}$ produced by the classifier ${g}$. ROIs are stochastically sampled instead of fixed ROIs~\cite{KolesnikovL16}. This has been shown to be more effective and avoids overfitting~\cite{negevsbelharbi2022}. Therefore, at each SGD step, a foreground pixel is sampled using a multinomial distribution over strong activations assuming that an object is local. Uniform distribution over low activations is used to sample background pixels assuming that background regions are evenly distributed in an image. The location of these two random pixels is encoded in ${\Omega^{\prime}_t }$. The partially pseudo-labeled mask for the sample ${\bm{X}_t}$ is denoted ${\bm{Y}_t}$, where ${\bm{Y}_t(p) \in \{0, 1\}^2}$ at location $p$ with value ${0}$ for background, ${1}$ for foreground,  and locations with unknown labels are encoded as unknown. At this stage, since the pre-trained classifier $f$ is frozen, its CAM ${\bm{C}_t}$ is fixed and does not change during the training of the decoder $f$, allowing for more stable sampling. The decoder output ${\bm{S}_t}$ is then aligned with ${\bm{Y}_t}$ using partial cross-entropy,
\begin{equation}
    \label{eq:pl}
    \begin{aligned}
    &\bm{H}_p(\bm{Y}_t, \bm{S}_t) = \\& - (1 - \bm{Y}_t(p))\; \log(\bm{S}_t^0(p))- \bm{Y}_t(p) \; \log(\bm{S}_t^1(p)) ,\; p \in \Omega\;.
    \end{aligned}
\end{equation}
More details on pixel-wise learning is provided in the appendix.

\noindent \textbf{b) Local consistency (CRF).} Standard CAMs have low resolution leading to a bloby effect where activations do not align with the object's boundaries, thus contributing to poor localization~\cite{choe2020evaluating,belharbi2022fcam}. To avoid this, a CRF loss~\cite{tang2018regularized} is employed to push the activations of ${\bm{S}_t}$ to be locally consistent in terms of pixels proximity and color. For a \emph{single image frame} $\bm{X}_t$ and its map ${\bm{S}_t}$, the CRF loss is formulated as,
\begin{equation}
    \label{eq:crf}
    \mathcal{R}(\bm{S}_t, \bm{X}_t) = \sum_{r \in \{0, 1\}} {\bm{S}^r_t}^{\top} \; \bm{W}_t \; (\bm{1} - \bm{S}^r_t) \;,
\end{equation}
where ${\bm{W}_t}$ is an affinity matrix  in which ${\bm{W}[i, j]}$ captures the \emph{proximity} between pair of pixels ${i, j}$ in the \emph{single frame} ${\bm{X}_t}$, and their \emph{color similarity}; ${r=0}$ is the index of the background, while ${r=1}$ is for the foreground. Since ${\bm{S}_t}$ has the same size as the input image, it is beneficial to minimize Eq.\ref{eq:crf}, since pixels' details is available, allowing for more accurate pixel-wise comparison.

\noindent \textbf{c) Absolute size constraint (ASC).} Unbalanced activations is a common issue in CAMs~\cite{choe2020evaluating,rony2023deep,belharbi2022fcam}. Often, strong activations cover only small and most discriminative parts of an object, allowing the background to dominate. Alternatively, large parts of an image are activated as foreground~\cite{rony2023deep}. To avoid such unbalance, global size prior over the CAM is employed. It consists of constraining the size of foreground and background regions to be as large as possible in a competitive way. To this end, Absolute Size Constraint (ASC)~\cite{belharbi2020minmaxuncer} on the decoder maps ${\bm{S}_t}$ is employed. This is achieved without requiring any knowledge of the true object size or prior knowledge of which region is larger. In particular, ASC allows the foreground and background region to compete by pushing their area to be as large as possible, effectively preventing unbalanced activation. This prior is formulated as inequality constraints which are then solved via a standard log-barrier method by minimizing the following loss,
\begin{equation}
\label{eq:log-bar-sz}
 \mathcal{R}_s(\bm{S}_t) = -\frac{1}{z} \left[ \log(\psi(\bm{S}^0_t)) + \log(\psi(\bm{S}^1_t)) \right] \;,
\end{equation}
where ${\psi(\bm{S}^0_t) = \sum \bm{S}^0_t}$ represents the area, \ie, size, of the background, and ${\psi(\bm{S}^0_t)}$ is the foreground area; ${z>0}$ is a weight that is increased periodically to strengthen the log-barrier loss.

\subsection{Multi-Frame Loss Based on Color Cues}
\label{subsec:multi-frameloss}
Our new temporal loss extends over a sequence of frames. It builds a fully connected graph over all pixel locations across all frames. Using discriminative response over the decoder's CAMs, and only the pairwise \emph{color similarity}, a CRF energy loss~\cite{tang2018regularized} is set to be minimized. This amounts to achieve co-localization across multiple frames using color appearance. The emerged object localization at frame level via per-frame terms are propagated through all the frames, and ensured to be consistent. Regions that have been deemed to be foreground are maintained as foreground across all frames, and the same way for background. Such knowledge transfer allows reducing errors caused by noisy pseudo-labels at frame level leading to better localization. Our approach is depicted in Fig.\ref{fig:proposal}. Our loss is based on the assumption that \emph{locally}, \ie over the extent of few frames, objects and overall frames maintain the same color, which is a fair assumption in video processing. Unlike TCAM~\cite{tcamsbelharbi2023}, we do not assume limited displacement of objects. This gives our method more flexibility to localize an object independently of its location.

Given a sequence of ${n}$ frames ${\bm{F}^{t, n}}$, we aim to consistently and simultaneously align their corresponding CAMs ${\bm{S}^{t, n}}$ with respect to \emph{color}. This is translated by \emph{explicitly} constraining the CAMs ${\bm{S}^{t, n}}$ to activate similarly on similar color pixels across all frames.
This is achieved by minimizing a color-only CRF loss~\cite{tang2018regularized} across all considered frames. We refer to the concatenation of frames in ${\bm{F}^{t, n}}$ as ${\bm{\bar{X}}}$. Similarly, we denote ${\bm{\bar{S}}}$ as the composite CAMs resulting from the concatenation of the corresponding CAMs ${\bm{M}^{t, n}}$ of the same frames ${\bm{X}^{t, n}}$. Concatenation can be done horizontally or vertically. Our temporal term loss is formulated as follows:
\begin{equation}
    \label{eq:crf_rgb}
    \mathcal{R}_c(\bm{F}^{t, n}, \bm{M}^{t, n}) = \sum_{r \in \{0, 1\}} \bm{\bar{S}}^{r\top} \; \bm{W} \; (\bm{1} - \bm{\bar{S}}^r) \;,
\end{equation}
where ${\bm{W}}$ is the \emph{color}-only similarity matrix between pixels of the concatenated image ${\bm{\bar{X}}}$.  
We note the main difference between CRF loss in Eq.\ref{eq:crf} which operates only on a single frame level, and accounts for both pairwise proximity and color similarity between pixels. On the other hand, our temporal CRF loss in Eq.\ref{eq:crf_rgb} operates on multiple frames, and acts only on pairwise color similarity across all frames without accounting for proximity between pixels or frames. This allows the object to be anywhere in the sequence or within a frame yielding more localization flexibility compared to TCAM method~\cite{tcamsbelharbi2023}.
In our implementation, we use a Gaussian kernel to capture color similarities. The kernel is implemented via the permutohedral lattice for fast computation. We refer to the Eq.\ref{eq:crf_rgb} as \emph{CoLoc} term. 

The loss in Eq.\ref{eq:crf_rgb} takes roughly ${\sim80}$ ms to process ${n=64}$ frames of size ${224\times224}$.  It is achieved via a hybrid implementation to speedup computation. The affinity matrix ${\bm{W}}$ is computed in parallel on CPU, while the matrix product is conducted over GPU. This yield a less computation overhead, and allows training in a practical time.

\subsection{Total Training Loss}
\label{subsec:total-loss}
Our final loss combines per-frame and multi-frame losses to be optimized. It is formulated as,
\begin{equation}
\label{eq:totalloss}
\begin{aligned}
\min_{\bm{\theta}} \quad & \sum_{p \in \Omega^{\prime}_t} \bm{H}_p(\bm{Y}_t, \bm{S}_t) + \lambda\; \mathcal{R}(\bm{S}_t, \bm{X}_t) + \mathcal{R}_s(\bm{S}_t)  
 \\ &  +  \frac{\lambda_c}{\abs{[\mathcal{R}_c]}}\; \mathcal{R}_c(\bm{F}^{t, n}, \bm{M}^{t, n}) \;,\\
\end{aligned}
\end{equation}
where $\lambda$ and $\lambda_c$ are positive weighting coefficients. All terms are optimized simultaneously through SGD method.

We note that the magnitude\footnote{Typical magnitude of ${\mathcal{R}_c}$ is ${2*10^9}$ for ${n=2}$, and can go up to ${2*10^{11}}$ for ${n=18}$. As a result, adequate values of the non-adapative ${\lambda_c}$ should be below ${2*10^{-9}}$.} of the term ${\mathcal{R}_c}$ increases with the number of frames $n$. Large magnitudes can easily overpower other terms in Eq.\ref{eq:totalloss}, hindering learning. In practice, this makes tuning the hyper-parameter ${\lambda_c}$ critical and challenging. To reduce this strong dependency on $n$ and stabilize learning, we propose an \emph{adaptive} weight ${\lambda_c}$ that automatically \emph{scales down} this term by its magnitude ${\abs{[\mathcal{R}_c]}}$. This is inspired by the recently proposed adaptive weight decay~\cite{Ghiasi2023} which uses the weight norm and its gradient for adaptation.
In this context, the operation ${[\cdot]}$ indicates that this value is now a constant and does not depend on any optimization parameters, \ie, ${\bm{\theta}}$. As a result, this term is always constant ${\lambda_c \; \frac{\mathcal{R}_c}{\abs{[\mathcal{R}_c]}} = \pm \lambda_c}$. However, its derivative is non-zero ${\partial \left( \lambda_c \; \frac{\mathcal{R}_c}{\abs{[\mathcal{R}_c]}} \right) / \partial \bm{\theta} = \frac{\lambda_c}{\abs{[\mathcal{R}_c]}} \frac{\partial \mathcal{R}_c}{\partial \bm{\theta}}}$. In this adaptive setup, the value ${\lambda_c}$ simply amplifies the scaled gradient. Its practical values are ${\lambda_c > 1}$, which makes it easier to tune compared to its non-adaptive version, which will be demonstrated empirically later. Most importantly, the adaptive version creates less dependency on $n$.

After being trained using Eq. \ref{eq:totalloss}, our model is applied over single frames using a forward pass for inference (Fig.\ref{fig:proposal-details}). This allows for fast inference and parallel computation over a video stream. Using standard procedures~\cite{tcamsbelharbi2023,choe2020evaluating}, a localization bounding box is extracted from the foreground CAM of the decoder ${\bm{S}^1_t}$. Our method yields a single model with parameters ${\{\bm{\theta^{\prime}}, \bm{\theta}\}}$ performing both tasks, classification and localization, while handling multiple classes at once.

\section{Results and Discussion}
\label{sec:results}

\subsection{Experimental Methodology}
\label{subsec:exp-method}

\noindent \textbf{Datasets.} To evaluate our method, we conducted experiments on two standard public unconstrained video datasets for WSVOL. For training, each video is labeled globally via a class. Frame bounding boxes are provided to evaluate localization. Following the literature, we used two challenging public datasets from YouTube\footnote{\url{https://www.youtube.com}}:  YouTube-Object v1.0 (\ytovone~\cite{prest2012learning}) and YouTube-Object v2.2 (\ytovtwodtwo~\cite{KalogeitonFS16} ). In all our experiments, we followed the same protocol as described in~\cite{prest2012learning,tcamsbelharbi2023,KalogeitonFS16}.

\noindent\textit{YouTube-Object v1.0 (\ytovone)~\cite{prest2012learning}:} A dataset composed of videos collected from YouTube by querying the names of 10 classes. A class has between 9-24 videos with a duration ranging from 30 seconds to 3 minutes. It has 155 videos, each split into short-duration clips, \ie, shots. There are 5507 shots with 571,089 frames in all. Images resolution vary from a video to another. It can go from ${400\times 300}$ up to ${1280\times 720}$.
The dataset has been divided by the authors~\cite{prest2012learning} into 27 test videos for a total of 396 bounding boxes, and 128 videos for training. Some videos from the train set are considered for validation to perform early-stopping~\cite{prest2012learning,tcamsbelharbi2023,KalogeitonFS16}.  We sampled 5 random videos per class to build a validation set with a total of 50 videos. Frame bounding box annotations are provided for test set.

\noindent\textit{YouTube-Object v2.2 (\ytovtwodtwo)~\cite{KalogeitonFS16}:} This is a large dataset built on \ytovone dataset with large and challenging test set. It is composed of more frames, 722,040 in total. The dataset was divided by the authors into 106 videos for training, and 49 videos for test. Following~\cite{tcamsbelharbi2023}, we sampled 3 random videos per class from the train set to build a validation set for model selection via early stopping. The test set has more samples compared to \ytovone. It is composed of a total of 2,667 bounding box making it a much more challenging dataset. We note that both datasets came already divided into train and test set by their authors. We use the same splits. Only the validation set used for model selection is selected randomly from the provided train set, as done in previous works.

\noindent \textbf{Implementation Details.} 
For all our experiments, we trained for 10 epochs with a mini-batch size of 32. ResNet50~\cite{heZRS16} was used as a backbone. Further ablations are done with VGG16~\cite{SimonyanZ14a}, InceptionV3~\cite{SzegedyVISW16}. Unless mentioned differently, all results are obtained using ResNet50 backbone. Images are resized to ${256\times256}$, and randomly cropped patches of size ${224\times224}$ are used for training. The temporal dependency ${n}$ in Eq.\ref{eq:crf_rgb} is set through the validation set from ${n \in \{2, \cdots, 18\}}$. The hyper-parameter ${\lambda_c}$ in Eq.\ref{eq:totalloss} is set from ${\{1, \cdots, 16\}}$ with the adaptive setup. In Eq.\ref{eq:crf_rgb}, frames are concatenated horizontally to build ${\bm{\bar{X}}}$.
We set the CRF hyper-parameter $\lambda$ in Eq.\ref{eq:totalloss} to ${2e^{-9}}$ -- the same value as in~\cite{tang2018regularized}.  Its color and spatial kernel bandwidth are set to 15 and 100, respectively, following~\cite{tang2018regularized}. The same color bandwidth is used for Eq.\ref{eq:crf_rgb}. The initial value of the log-barrier coefficient, $z$ in Eq.\ref{eq:log-bar-sz}, is set to $1$ and then increases by a factor of ${1.01}$ in each epoch with a maximum value of $10$. The best learning rate was selected within ${\{0.1, 0.01, 0.001\}}$. The classifier $g$ was pre-trained on single frames on the train set. 

\noindent \textbf{Baseline Methods.} For comparison, publicly available results were considered. We compared our proposed approach with different WSVOL methods~\cite{Croitoru2019,Halle2017,joulin2014,Kwak2015,papazoglou2013fast,prest2012learning,Tokmakov2016,tsai2016semantic}, POD~\cite{jun2016pod}, SPFTN~\cite{zhang2020spftn}, and FPPVOS~\cite{umer2021efficient}. We also considered the performance of several CAM-based methods reported in~\cite{tcamsbelharbi2023}: CAM~\cite{zhou2016learning}, GradCAM~\cite{SelvarajuCDVPB17iccvgradcam}, GradCam++~\cite{ChattopadhyaySH18wacvgradcampp}, Smooth-GradCAM++~\cite{omeiza2019corr}, XGradCAM~\cite{fu2020axiom}, and LayerCAM~\cite{JiangZHCW21layercam}, which were all trained on single frames, \ie, no temporal dependency was considered. We compare our results as well to CLIP-ES~\cite{lin2023} using true class as input, and without background list or CAM-refinement for a fair comparison. The LayerCAM method~\cite{JiangZHCW21layercam} was employed to generate classifier CAMs ${\bm{C}_t}$ used to create pseudo-labels (${\bm{Y}_t}$) for our method (Eq.\ref{eq:pl}). Note that different CAM-based methods can be integrated with our approach.

\noindent \textbf{Evaluation Metric.} Localization accuracy was evaluated using the \corloc metric~\cite{tcamsbelharbi2023}. It describes the percentage of predicted bounding boxes that have an Intersection Over Union (IoU) between prediction and ground truth greater than half (IoU ${>50\%}$). In addition, the  inference time per frame of size ${224\times224}$ is reported over GPU (NVIDIA Tesla P100) and CPU (Intel(R) Xeon(R), 48 cores). The same hardware is used to report the computation time of our multi-frame loss (Eq.\ref{eq:crf_rgb}).

\begin{table*}
\begin{center}
\resizebox{\linewidth}{!}{
\begin{tabular}{|l||*{10}{c|}|g|c|}
\hline
\multicolumn{13}{|c|}{\textbf{\ytovone dataset}} \\
\hline
\textbf{Method {\small \emph{(venue, publication year)}}} & \textbf{Aero} & \textbf{Bird} & \textbf{Boat} & \textbf{Car} & \textbf{Cat} & \textbf{Cow} & \textbf{Dog} & \textbf{Horse} & \textbf{Mbike} & \textbf{Train} & \textbf{Avg} & \textbf{Time/Frame} \\
\hhline{----------||---}
\hhline{----------||---}
Prest et al.~\cite{prest2012learning} {\small \emph{(cvpr, 2012)}} & 51.7 & 17.5 & 34.4 & 34.7 & 22.3 & 17.9 & 13.5 & 26.7 & 41.2 & 25.0 & 28.5 & N/A   \\
Papazoglou et al.~\cite{papazoglou2013fast} {\small \emph{(iccv, 2013)}} & 65.4 & 67.3 & 38.9 & 65.2 & 46.3 & 40.2 & 65.3 & 48.4 & 39.0 & 25.0 & 50.1 & 4s  \\
Joulin et al.~\cite{joulin2014} {\small \emph{(eccv, 2014)}} & 25.1 & 31.2 & 27.8 & 38.5 & 41.2 & 28.4 & 33.9 & 35.6 & 23.1 & 25.0 & 31.0 & N/A \\
Kwak et al.~\cite{Kwak2015} {\small \emph{(iccv, 2015)}}  & 56.5 & 66.4 & 58.0 & 76.8 & 39.9 & 69.3 & 50.4 & 56.3 & 53.0 & 31.0 & 55.7 & N/A  \\
Rochan et al.~\cite{RochanRBW16} {\small \emph{(ivc, 2016)}} & 60.8 & 54.6 & 34.7 & 57.4 & 19.2 & 42.1 & 35.8 & 30.4 & 11.7 & 11.4 & 35.8 & N/A \\
Tokmakov et al.~\cite{Tokmakov2016} {\small \emph{(eccv, 2016)}} & 71.5 & 74.0 & 44.8 & 72.3 & 52.0 & 46.4 & 71.9 & 54.6 & 45.9 & 32.1 & 56.6 & N/A  \\
POD~\cite{jun2016pod} {\small \emph{(cvpr, 2016)}} & 64.3 & 63.2 & 73.3 & 68.9 & 44.4 & 62.5 & 71.4 & 52.3 & 78.6 & 23.1 & 60.2 & N/A  \\
Tsai et al.~\cite{tsai2016semantic} {\small \emph{(eccv, 2016)}} & 66.1 & 59.8 & 63.1 & 72.5 & 54.0 & 64.9 & 66.2 & 50.6 & 39.3 & 42.5 & 57.9 & N/A  \\
Haller et al.~\cite{Halle2017} {\small \emph{(iccv, 2017)}} & 76.3 & 71.4 & 65.0 & 58.9 & 68.0 & 55.9 & 70.6 & 33.3 & 69.7 & 42.4 & 61.1 & 0.35s  \\
Croitoru et al.~\cite{Croitoru2019} (LR-N\textsubscript{iter1}) {\small \emph{(ijcv, 2019)}} & 77.0 & 67.5 & 77.2 & 68.4 & 54.5 & 68.3 & 72.0 & 56.7 & 44.1 & 34.9 & 62.1 & 0.02s  \\
Croitoru et al.~\cite{Croitoru2019} (LR-N\textsubscript{iter2}) {\small \emph{(ijcv, 2019)}} & 79.7 & 67.5 & 68.3 & 69.6 & 59.4 & 75.0 & 78.7 & 48.3 & 48.5 & 39.5 & 63.5 & 0.02s\\
Croitoru et al.~\cite{Croitoru2019} (DU-N\textsubscript{iter2}) {\small \emph{(ijcv, 2019)}} & 85.1 & 72.7 & 76.2 & 68.4 & 59.4 & 76.7 & 77.3 & 46.7 & 48.5 & 46.5 & 65.8 & 0.02s \\
Croitoru et al.~\cite{Croitoru2019} (MS-N\textsubscript{iter2}) {\small \emph{(ijcv, 2019)}} & 84.7 & 72.7 & 78.2 & 69.6 & 60.4 & 80.0 & 78.7 & 51.7 & 50.0 & 46.5 & 67.3 & 0.15s \\
SPFTN (M)~\cite{zhang2020spftn} {\small \emph{(tpami, 2020)}} & 66.4 & 73.8 & 63.3 & 83.4 & 54.5 & 58.9 & 61.3 & 45.4 & 55.5 & 30.1 & 59.3 & N/A  \\
SPFTN (P)~\cite{zhang2020spftn} {\small \emph{(tpami, 2020)}}& \textbf{97.3} & 27.8 & 81.1 & 65.1 & 56.6 & 72.5 & 59.5 & \textbf{81.8} & 79.4 & 22.1 & 64.3 & N/A  \\
FPPVOS~\cite{umer2021efficient} {\small \emph{(optik, 2021)}} & 77.0 & 72.3 & 64.7 & 67.4 & 79.2 & 58.3 & 74.7 & 45.2 & 80.4 & 42.6 & 65.8 & 0.29s  \\
\cline{1-13}
CAM~\cite{zhou2016learning} {\small \emph{(cvpr, 2016)}} & 75.0 & 55.5 & 43.2 & 69.7 & 33.3 & 52.4 & 32.4 & 74.2 & 14.8 & 50.0 & 50.1 & 0.2ms  \\
GradCAM~\cite{SelvarajuCDVPB17iccvgradcam} {\small \emph{(iccv, 2017)}} & 86.9 & 63.0 & 51.3 & 81.8 & 45.4 & 62.0 & 37.8 & 67.7 & 18.5 & 50.0 & 56.4 & 27.8ms  \\
GradCAM++~\cite{ChattopadhyaySH18wacvgradcampp} {\small \emph{(wacv,2018)}} & 79.8 & 85.1 & 37.8 & 81.8 & 75.7 & 52.4 & 64.9 & 64.5 & 33.3 & 56.2 & 63.2 & 28.0ms  \\
Smooth-GradCAM++~\cite{omeiza2019corr} {\small \emph{(corr, 2019)}} & 78.6 & 59.2 & 56.7 & 60.6 & 42.4 & 61.9 & 56.7 & 64.5 & 40.7 & 50.0 & 57.1 & 136.2ms  \\
XGradCAM~\cite{fu2020axiom} {\small \emph{(bmvc, 2020)}} & 79.8 & 70.4 & 54.0 & 87.8 & 33.3 & 52.4 & 37.8 & 64.5 & 37.0 & 50.0 & 56.7 & 14.2ms  \\
LayerCAM~\cite{JiangZHCW21layercam} {\small \emph{(ieee, 2021)}} & 85.7 & \textbf{88.9} & 45.9 & 78.8 & 75.5 & 61.9 & 64.9 & 64.5 & 33.3 & 56.2 & 65.6 & 17.9ms  \\
F-CAM~\cite{belharbi2022fcam} {\small \emph{(wacv, 2022)}} & 75.4 & 85.7 & 75.6 & 85.4 & 59.2 & 87.0 & 52.9 & 66.6 & 62.5 & 50.0 & 70.0 & 18.5ms  \\
CLIP-ES~\cite{lin2023} {\small \emph{(cvpr, 2023)}}  & 46.4 & 74.0 & 5.4 & \textbf{90.9} & \textbf{90.9} & 66.6 & 75.6 & 83.8 & 74.0 & 43.7 & 65.1 & 49.9ms  \\
TCAM~\cite{tcamsbelharbi2023} with Layer-CAM {\small \emph{(wacv, 2023)}} & 90.5 & 70.4 & 62.2 & 75.7 & 84.8 & \textbf{81.0} & 81.0 & 64.5 & 70.4 & 50.0 & 73.0 & 18.5ms  \\
\colocam with Layer-CAM (ours) & 90.4 & 74.0 & \textbf{91.8} & 87.8 & 78.7 & 80.9 & \textbf{89.1} & 74.1 & \textbf{85.1} & \textbf{68.7} & \textbf{82.1} & 18.5ms  \\
\hline
\hline
\multicolumn{13}{|c|}{\textbf{\ytovtwodtwo dataset}} \\
\hline
\textbf{Method {\small \emph{(venue, publication year)}}} & \textbf{Aero} & \textbf{Bird} & \textbf{Boat} & \textbf{Car} & \textbf{Cat} & \textbf{Cow} & \textbf{Dog} & \textbf{Horse} & \textbf{Mbike} & \textbf{Train} & \textbf{Avg} & \textbf{Time/Frame} \\
\hhline{----------||---}
\hhline{----------||---}
Haller et al.~\cite{Halle2017}  {\small \emph{(iccv, 2017)}}& 76.3 & 68.5 & 54.5 & 50.4 & 59.8 & 42.4 & 53.5 & 30.0 & 53.5 & \textbf{60.7} & 54.9 & 0.35s   \\
Croitoru et al.~\cite{Croitoru2019} (LR-N\textsubscript{iter1}) {\small \emph{(ijcv, 2019)}} & 75.7 & 56.0 & 52.7 & 57.3 & 46.9 & 57.0 & 48.9 & 44.0 & 27.2 & 56.2 & 52.2 & 0.02s    \\
Croitoru et al.~\cite{Croitoru2019} (LR-N\textsubscript{iter2}) {\small \emph{(ijcv, 2019)}} & 78.1 & 51.8 & 49.0 & 60.5 & 44.8 & 62.3 & 52.9 & 48.9 & 30.6 & 54.6 &  53.4 & 0.02s    \\
Croitoru et al.~\cite{Croitoru2019} (DU-N\textsubscript{iter2}){\small \emph{(ijcv, 2019)}} & 74.9 & 50.7 & 50.7 & 60.9 & 45.7 & 60.1 & 54.4 & 42.9 & 30.6 & 57.8 & 52.9 & 0.02s   \\
Croitoru et al.~\cite{Croitoru2019} (BU-N\textsubscript{iter2}){\small \emph{(ijcv, 2019)}} & 82.2 & 51.8 & 51.5 & 62.0 & 50.9 & 64.8 & 55.5 & 45.7 & 35.3 & 55.9 & 55.6 & 0.02s  \\
Croitoru et al.~\cite{Croitoru2019} (MS-N\textsubscript{iter2}){\small \emph{(ijcv, 2019)}} & 81.7 & 51.5 & 54.1 & 62.5 & 49.7 & 68.8 & 55.9 & 50.4 & 33.3 & 57.0 & 56.5 & 0.15s  \\
\cline{1-13}
CAM~\cite{zhou2016learning} {\small \emph{(cvpr, 2016)}} & 52.3 & 66.4 & 25.0 & 66.4 & 39.7 & \textbf{87.8} & 34.7 & 53.6 & 45.4 & 43.7 & 51.5 & 0.2ms  \\
GradCAM~\cite{SelvarajuCDVPB17iccvgradcam} {\small \emph{(iccv, 2017)}} & 44.1 & 68.4 & 50.0 & 61.1 & 51.8 & 79.3 & 56.0 & 47.0 & 44.8 & 42.4 & 54.5 & 27.8ms  \\
GradCAM++~\cite{ChattopadhyaySH18wacvgradcampp} {\small \emph{(wacv, 2018)}} & 74.7 & 78.1 & 38.2 & 69.7 & 56.7 & 84.3 & 61.6 & 61.9 & 43.0 & 44.3 & 61.2 & 28.0ms  \\
Smooth-GradCAM++~\cite{omeiza2019corr} {\small \emph{(corr, 2019)}} & 74.1 & 83.2 & 38.2 & 64.2 & 49.6 & 82.1 & 57.3 & 52.0 & 51.1 & 42.4 & 59.5 & 136.2ms  \\
XGradCAM~\cite{fu2020axiom} {\small \emph{(bmvc, 2020)}} & 68.2 & 44.5 & 45.8 & 64.0 & 46.8 & 86.4 & 44.0 & 57.0 & 44.9 & 45.0 & 54.6 & 14.2ms  \\
LayerCAM~\cite{JiangZHCW21layercam} {\small \emph{(ieee, 2021)}} & 80.0 & 84.5 & 47.2 & 73.5 & 55.3 & 83.6 & 71.3 & 60.8 & 55.7 & 48.1 & 66.0 & 17.9ms  \\
F-CAM~\cite{belharbi2022fcam} {\small \emph{(wacv, 2022)}} & 77.1 & 89.2 & 73.1 & \textbf{81.8} & 51.8 & 80.6 & 47.0 & 62.9 & 62.5 & 50.0 & 67.6 & 18.5ms  \\
CLIP-ES~\cite{lin2023} {\small \emph{(cvpr, 2023)}}  & 51.7 & 78.7 & 11.8 & 63.9 & 75.8 & 78.5 & 77.4 & 64.6 & 54.5 & 45.5 & 60.2 & 49.9ms  \\
TCAM~\cite{tcamsbelharbi2023} with Layer-CAM {\small \emph{(wacv, 2023)}}  & 79.4 & \textbf{94.9} & 75.7 & 61.7 & 68.8 & 87.1 & 75.0 & 62.4 & 72.1 & 45.0 & 72.2 & 18.5ms  \\
\colocam with Layer-CAM (ours) & \textbf{82.9} & 92.2 & \textbf{85.4} & 67.7 & \textbf{80.1} & 85.7 & \textbf{79.2} & \textbf{67.4} & \textbf{72.7} & 58.2 & \textbf{77.1} & 18.5ms  \\
\hline
\end{tabular}}
\end{center}
\caption{\corloc localization accuracy (\%) per-class and average over classes (Avg) of \colocam and state-of-the-art methods on the \ytovone~\cite{prest2012learning} and \ytovtwodtwo~\cite{KalogeitonFS16} test sets when using the ResNet50 CNN backbone. The computation inference time is over a single image with a size of ${224\times224}$.}
\label{tab:yto}
\end{table*}

\subsection{Comparison with the State-of-the-Art}
\label{subsec:compare}

The results in Tab.\ref{tab:yto} show that our \colocam method outperforms other methods by a large margin, more so on \ytovone compared to \ytovtwodtwo. This highlights the difficulty of \ytovtwodtwo. In term of per-class performance, our method is competitive in most classes. In particular, we observe that our method achieved a large improvement over the challenging class 'Train'. In general, our method and CAM methods are still behind on the class 'Horse' and 'Aero' compared to the SPFTN method~\cite{zhang2020spftn} which relies on optical flow to generate proposals. Both classes present different challenges. The former class shows with dense multi-instances, while the latter, \ie, 'Aero', appears in complex scenes (airport), often with very large size, and occlusion. In term of inference time, our method is relatively fast which is similar to TCAM~\cite{tcamsbelharbi2023} and F-CAM~\cite{belharbi2022fcam}.

\begin{figure*}[ht!]
     \centering
     \begin{subfigure}[b]{0.46\linewidth}
         \centering
         \includegraphics[width=\linewidth]{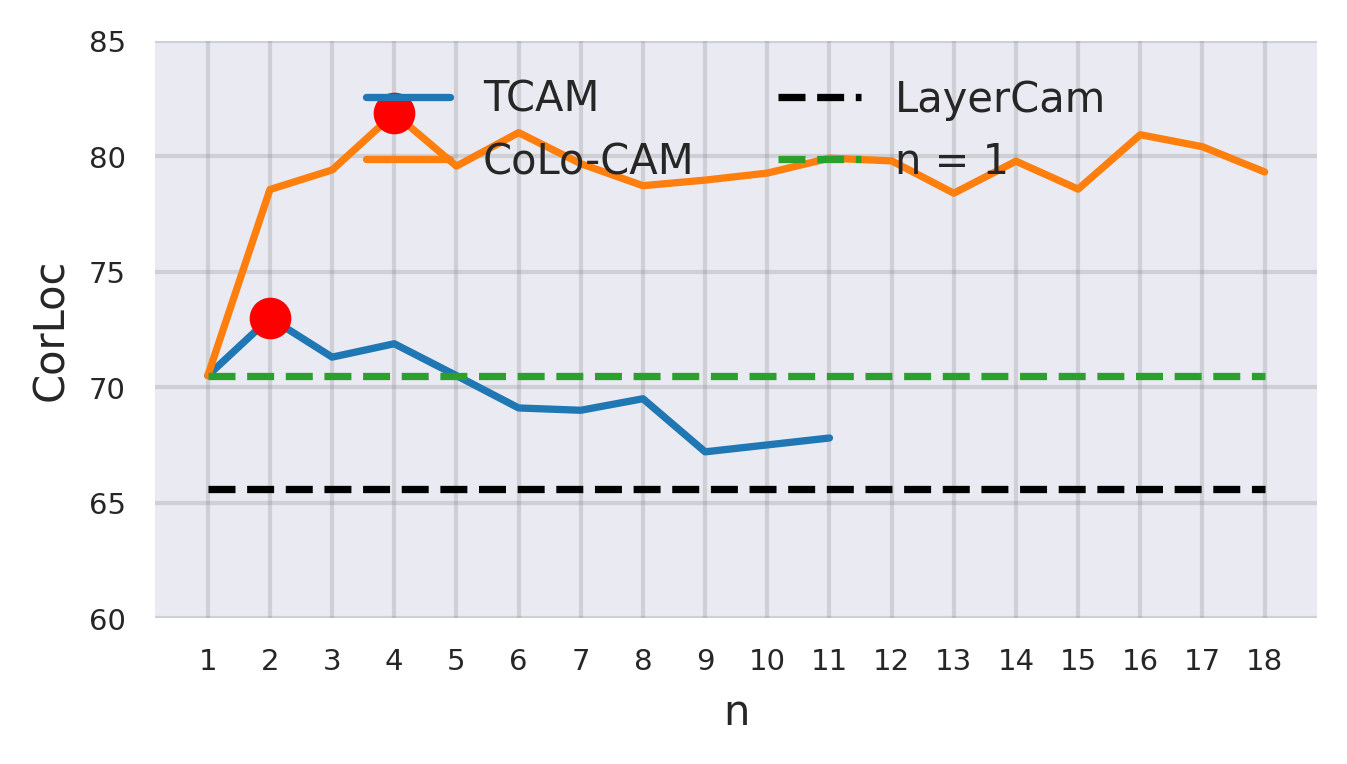}
         \caption{}
         \label{fig:ablation-range-time}
     \end{subfigure}
     \begin{subfigure}[b]{0.46\linewidth}
         \centering
         \includegraphics[width=\linewidth]{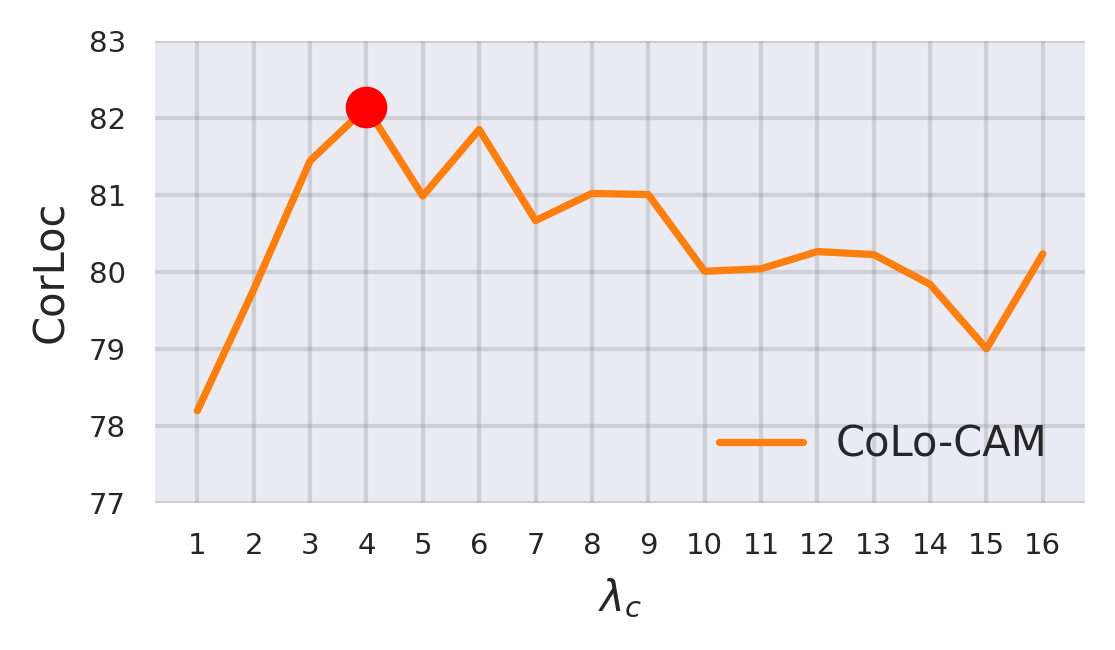}
         \caption{}
         \label{fig:ablation-lambda-c}
     \end{subfigure}
     \\
     \begin{subfigure}[b]{0.49\linewidth}
         \centering
         \includegraphics[width=\linewidth]{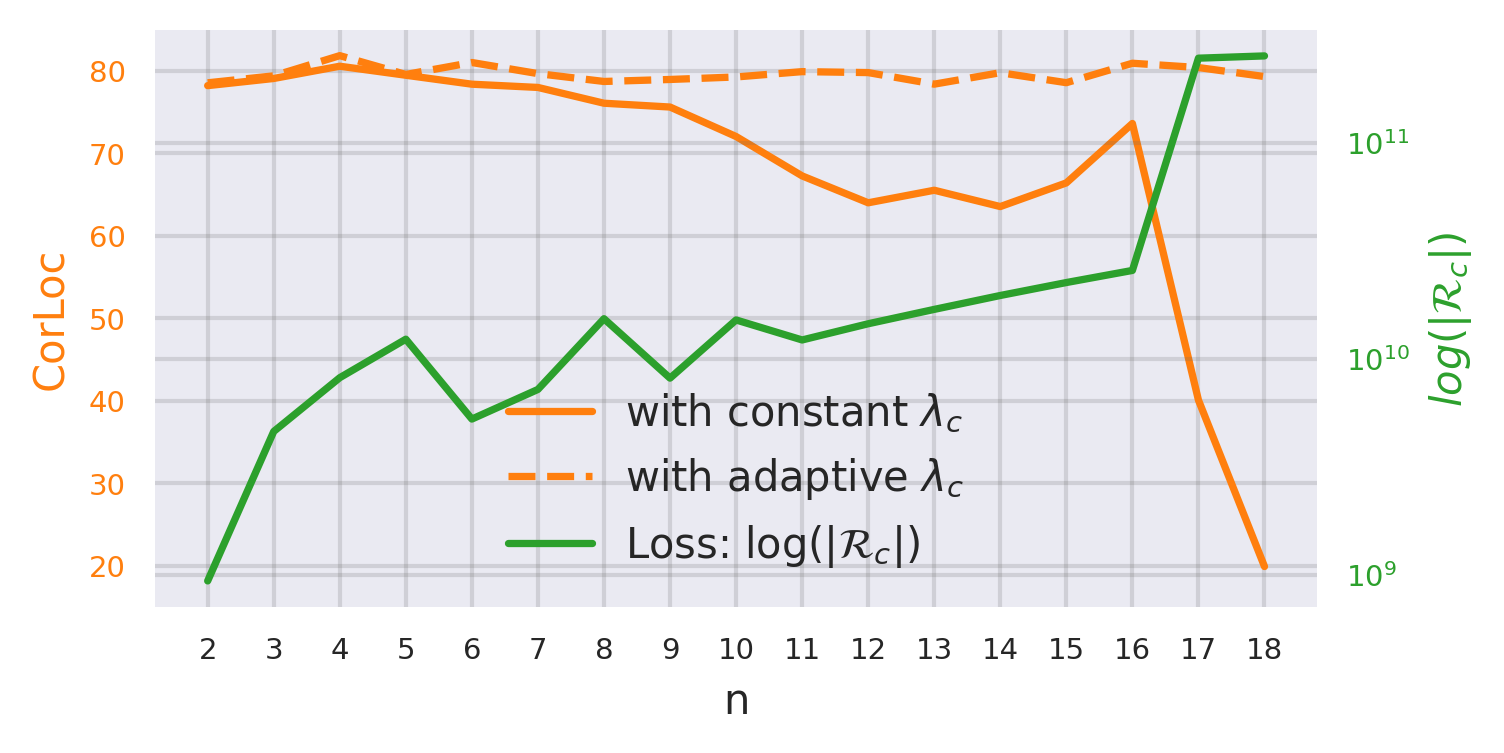}
         \caption{}
         \label{fig:ablation-lambda-c-const-vs-adap}
     \end{subfigure}
     \begin{subfigure}[b]{0.46\linewidth}
         \centering
         \includegraphics[width=\linewidth]{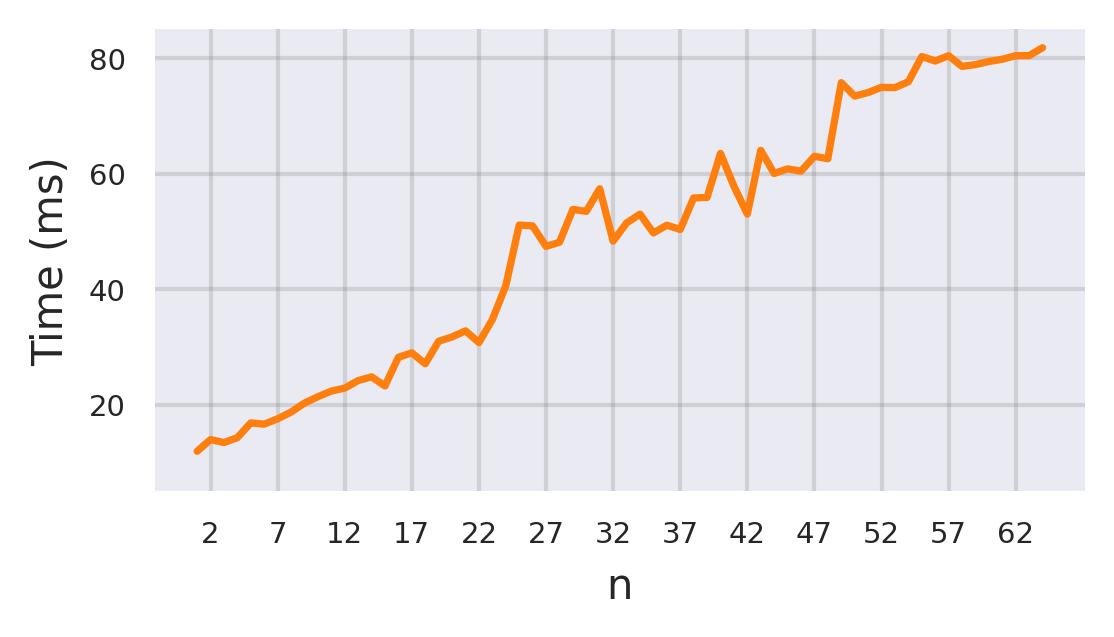}
         \caption{}
     \label{fig:ablation-time-joint-crf}
     \end{subfigure}
        \caption{Ablations on the \ytovone test set. 
        \textbf{(a)} Impact of the number of frames ${n}$ over \corloc accuracy. 
        \textbf{(b)} Impact of the \emph{adaptive} ${\lambda_c}$ over \corloc accuracy.
        \textbf{(c)} \emph{Left y-axis, Orange}: \corloc accuracy on the \ytovone test set using constant and adaptive ${\lambda_c}$ weight. \emph{Right y-axis, Green}: ${\log{(\abs{\mathcal{R}_c})}}$ . The \emph{x-axis} is the number of frames $n$. (Better visualized in color)
        \textbf{(d)} Computation time of our multi-frame loss term (Eq.\ref{eq:crf_rgb}) in function of number of frames $n$. 
        }
        \label{fig:ablations-lambda-c-n}
\end{figure*}
\subsection{Ablation Studies}

The following experiments provide insight on how localization performance is affected by loss terms, backbone, sampling strategy. Computation time of our multi-frame loss is reported as well in addition to performance stability of our method.

{
\begin{table}[ht!]
\centering
\resizebox{\linewidth}{!}{%
\centering
\begin{tabular}{|lllc|c|}
\hline
\multicolumn{3}{|l}{\textbf{Methods}}& &  \corloc (\%) \\
\hline \hline 
\multicolumn{3}{|l}{Layer-CAM~\cite{JiangZHCW21layercam} {\small \emph{(ieee)}}} &  &  65.6  \\
\hline 
\multirow{3}{*}{Single-frame} & &       
PLs          &  &   $68.5$       \\
&& PLs + CRF &  &   $69.6$       \\
&& PLs + ASC &  &   $66.2$      \\
&& PLs + ASC + CRF &  &   $70.0$\\ 
\hline 
Multi-frame & & PLs + ASC + CRF + CoLoc (Ours) &  &   \textbf{82.1} \\
\hline
\multicolumn{3}{|l}{Improvement} &  &  +$16.5$ \%\\ \hline 
\end{tabular}
}
\caption{Impact of different \colocam loss terms on \corloc localization accuracy. Results are reported on the \ytovone test set. All the cases use the same low resolution CAMs built from a pretrained Layer-CAM~\cite{JiangZHCW21layercam} model.}
\label{tab:ablation-parts}
\vspace{-1em}
\end{table}
}

{
\begin{table}[ht!]
\centering
\resizebox{\linewidth}{!}{%
\centering
\begin{tabular}{|lllc|c|c|}
\hline
\multicolumn{3}{|l}{\textbf{Methods}}& &  \ytovone & \ytovtwodtwo  \\
\hline \hline 
\multirow{1}{*}{Single-frame} & &       
Layer-CAM~\cite{JiangZHCW21layercam}          &  &   $65.6$      & $66.0$ \\
&& F-CAM~\cite{belharbi2022fcam}              &  &   $70.0$      & $67.6$ \\
\hline 
\multirow{1}{*}{Multi-frame} & &       
TCAM~\cite{tcamsbelharbi2023} (F-CAM + temporal max-pooling)          &  &   $73.0$    & $72.2$   \\
&& \colocam (F-CAM + our temporal module)                             &  &   \bm{$82.1$}    & \bm{$77.1$}   \\
\hline
\end{tabular}
}
\caption{Comparison of F-CAM~\cite{belharbi2022fcam}, TCAM~\cite{tcamsbelharbi2023}, and our proposed \colocam in term of \corloc localization accuracy on the \ytovone and \ytovtwodtwo test sets. All the methods use the same low resolution CAMs built from a pretrained Layer-CAM~\cite{JiangZHCW21layercam} model.}
\label{tab:tcam-fcam-vs-ours}
\vspace{-1em}
\end{table}
}

\noindent\textbf{a) Impact of different loss terms (Tab.\ref{tab:ablation-parts}).} 
Without any spatiotemporal dependency, the use of pseudo-labels, CRF, and the absolute size constraint (ASC) helped to improve the localization performance (Tab.\ref{tab:ablation-parts}). This brought about the localization performance from ${65.6\%}$ to ${70.0\%}$. However, adding our multi-frame term, \ie CoLoc, increased the performance up to ${82.1\%}$ demonstrating its benefit. Furthermore, Tab.\ref{tab:tcam-fcam-vs-ours} provides a comparison between F-CAM~\cite{belharbi2022fcam}, TCAM~\cite{tcamsbelharbi2023}, and our method. Note that F-CAM, TCAM and our method use the same per-frame losses, same architecture, and same low resolution CAMs, estimated from Layer-CAM~\cite{JiangZHCW21layercam} pretrained model, for a fair comparison. These results confirm the benefit of our new temporal loss.

\noindent\textbf{b) Impact of ${n}$ on localization performance (Fig.\ref{fig:ablation-range-time}).} 
We observe that both methods, TCAM~\cite{tcamsbelharbi2023} and ours, provide better results when considering spatiotemporal information. However, TCAM reached its peak performance when using only ${n=2}$ frames. Large performance degradation is observed when increasing $n$. This comes as a result of assuming that objects have limited movement. Their proposed ROIs collector can mistakenly gather a true ROI in a distant frame, but, spatially, it is no longer an ROI but background in the current frame because the object has moved.
On the opposite, our method improves when increasing $n$ until it reaches its peak at ${n=4}$. Different from TCAM, our method showed more robustness and stability with the increase of $n$ since we only assume the object's visual similarity. When using large time dependency (${n=18}$), our method still maintains a high localization performance (${\sim80\%}$) demonstrating its robustness against long dependency.

\noindent\textbf{c) Constant vs adaptive ${\lambda_c}$ (Fig.\ref{fig:ablation-lambda-c-const-vs-adap}).}  
Note that in the constant setup, adequate ${\lambda_c}$ can be determined using a grid search or given a prior on the magnitude loss. Both approaches are tedious, especially when ideal values of ${\lambda_c}$ are below ${2*10^{-9}}$ creating a large search space. Exploring such space is computationally expensive. Using constant value provides good results up to 6 frames, while larger $n$ values lead to performance degradation, as adequate values are required (Fig.\ref{fig:ablation-lambda-c-const-vs-adap}). Our adaptive scheme is more intuitive and does not require prior knowledge. It achieved constantly better and more stable performance with less effort. Better results are obtained with ${\lambda_c}$ values between 3 and 7 (Fig.\ref{fig:ablation-lambda-c}). Performance degradation is observed when using large amplification, giving more importance to the ${\mathcal{R}_c}$ term, allowing it to outweigh other terms which hindered training. In contrast, using small amplification, \eg ${\mathcal{R}_c \in \{1, 2}\}$, gave a small push to this term, hence, less contribution.

\noindent\textbf{d) Computation time of our multi-frame loss (Eq. \ref{eq:crf_rgb}) with respect to ${n}$ (Fig.\ref{fig:ablation-time-joint-crf}).}
The computation of our multi-frame loss term in Eq. \ref{eq:crf_rgb} is split over CPU and GPU devices to perform hybrid computation. The left-side matrix product is performed on GPU, while the rest of the term is computed on CPU. Note that the transfer time from CPU to GPU is included in our time analysis. As shown in Fig.\ref{fig:ablation-time-joint-crf}, computation overhead in processing time grows linearly with respect to the number of frames ${n}$. However, this computational cost remains reasonable for training in a realistic time, even with a large $n$. For example, $n = 64$ frames can be processed in ${\sim 81}$ ms. Furthermore, full training of our method over \ytovone for 10 epochs with ${n=4}$ takes less than an hour (${~\sim45}$ mins).

We conducted furthermore analysis of the computational cost of our method during training and inference, while comparing to its baseline Layer-CAM method~\cite{JiangZHCW21layercam}. Table.\ref{tab:compl-time} shows the obtained results.
Over training, the backward pass with all the losses and parameters update of our method have a high cost amounting to 0.84s for single SGD step while forward pass of ${n=64}$ reaches ${\sim80}$. Its counterpart method achieves 0.06s. However, this training speed of our method remains practical.
At inference time, our method is relatively faster than Layer-CAM since this last one requires gradient computation. For instance, our method can process image with a rate of 464 FPS compared to Layer-CAM rate of 128 FPS which over-loads the GPU memory.

Although, multiframe loss does not add a large computational overhead, it can be further sped up. One way to achieve this is to reduce the number of nodes in the fully connected graph. This can be done by considering a node as a small region such as an image superpixel or a patch. However, this required an adapted heuristic to describe the color of each region and reduce it to a single value similar to a pixel, along with the regions' corresponding probabilities. Another way is to set the decoder to produce smaller localization maps than the full image size. This will effectively reduce the number of locations. In this case, the image can be simply interpolated to a small size.

{
\begin{table*}[hpt!]
\centering
\resizebox{0.7\linewidth}{!}{%
\centering
\begin{tabular}{lc||cc}
\hline
\textbf{Case/Approach} &&  Layer-CAM (w. ResNet50) & Ours (w. ResNet50) \\
\hline
\makecell[l]{Train time  1 SGD step \\
Batch size: 32}                            && 0.06s   & 0.84s  \\
\hline
Inference time per-frame                   && 19.99ms & 10.11ms  \\
\hline
\makecell[l]{Inference time: Frame rate \\
(FPS)}                                     && 128*    & 464  \\
\hline
Total n. params.                           &&  23.5M  & 32.5M \\
\hline
N. learnable params.                       &&  23.5M  &  9.0M   \\
\hline
N. FLOPs (GFLOPs) (1 frame)                && 38.23      &  55.49   \\
\hline
N. MACs  (GMACs)  (1 frame)                &&  19.07    &  27.70  \\
\hline
\end{tabular}
}
\caption{Comparison of computation time, number of parameters, number of FLOPs/MACs of basic model vs ours. Computations are performed on an NVIDIA A100. Frame size: ${224\times 224}$.
\newline
*: The frame rate (FPS) for Layer-CAM runs to an out of memory issue when going beyond 128 frames over a GPU with 40GB of memory. Processing 128 frames takes roughly 200ms. The FLOPs/MACs are computed using the library \href{https://pypi.org/project/calflops}{pypi.org/project/calflops} (V0.3.2).}
\label{tab:compl-time}
\end{table*}
}

\noindent\textbf{e) Effect of frame sampling strategy (Tab. \ref{tab:ablation-frame-sampling}).}
We have explored different strategies to randomly sample $n$ training frames from a video. We focus on \emph{frame diversity} to asses its impact on co-localization versus when frames are adjacent. We present three sampling techniques:

\noindent\emph{1) Adjacent scheme}: A first frame is uniformly sampled from the video. In addition, its previous ${n-1}$ frames are considered. This favors similar (less diverse) frames.

\noindent\emph{2) Interval scheme}: This is the opposite case of the adjacent scheme where the aim is to sample the most diverse frames. The video is divided into ${n}$ equal and disjoint intervals. Then, a single frame is uniformly sampled from each interval.

\noindent\emph{3) Gaussian scheme}: This is a middle-ground scheme which lies between adjacent and interval strategies. We sample $n$ random frames via a Gaussian distribution centered in the middle of a video.

All the sampling is done without repetition. Results in Tab.\ref{tab:ablation-frame-sampling} show that adjacent sampling performs the best, suggesting that our co-localization term is more beneficial when frames are relatively similar. This is consistent with the color assumption. Sampling diverse frames can lead to contrast change and/or major change in context.  Note that using diverse frames via the Gaussian or interval scheme still yields good performance. Unless mentioned otherwise, all our results are obtained via a adjacent scheme sampling.

{
\begin{table}[ht!]
\centering
\resizebox{0.7\linewidth}{!}{%
\centering
\begin{tabular}{|lc|c|}
\hline
\multicolumn{1}{|l}{\textbf{Methods}}& &  \corloc accuracy (\%)  \\
\hline \hline 
Adjacent     &  &   $\bm{82.1}$       \\
Gaussian        &  &   $80.2$       \\
Interval        &  &   $79.8$      \\
\hline 
\end{tabular}
}
\caption{Impact of frame sampling strategies on localization performance (\corloc) on \ytovone test set.}
\label{tab:ablation-frame-sampling}
\vspace{-1em}
\end{table}
}

{
\begin{table}[ht!]
\centering
\resizebox{\linewidth}{!}{%
\centering
\begin{tabular}{|lc|ccc|}
\hline
\multirow{2}{*}{Cases / architectures} & &  \multicolumn{3}{c|}{\corloc accuracy (\%)}  \\
\cline{3-5}
& &  ResNet50 & VGG16 & Inception  \\
\hline
Layer-CAM~\cite{JiangZHCW21layercam}         &  &  65.6    & 65.2 & 65.0 \\
\hline 
\makecell{\colocam with\\ Layer-CAM~\cite{JiangZHCW21layercam} (Ours)} &  &  \textbf{82.1}    & \textbf{73.9} & \textbf{76.9} \\
\hline
\end{tabular}
}
\caption{Impact of the classifier backbone over localization performance (\corloc) on \ytovone test set.}
\label{tab:ablation-arch}
\vspace{-1em}
\end{table}
}

\noindent\textbf{f) Effect of the classifier backbone architecture (Tab. \ref{tab:ablation-arch}).}
We studied the effect of the backbone classifier for our method over localization performance. To this end, we used three different common backbones for WSOL task~\cite{choe2020evaluating}: ResNet50~\cite{heZRS16}, VGG16~\cite{SimonyanZ14a}, and InceptionV3~\cite{SzegedyVISW16}. From Tab.\ref{tab:ablation-arch}, we observe that in all cases, our method improves the baseline and still outperforms state-of-the-art of ${73.0\%}$. However, there is a discrepancy in performance between the three backbones. Such difference is common in WSOL tasks in natural scene images~\cite{choe2020evaluating,belharbi2022fcam} and medical images~\cite{rony2023deep} where ResNet50 performs better. Furthermore, even though these backbones have similar performance on the test set using the baseline method (Layer-CAM~\cite{JiangZHCW21layercam}), they do not necessarily have the same performance on the train set used to train our method. Such a suspected difference could also explain the discrepancy in test results. Note that the train sets do not have bounding box annotations, which prevents measuring the localization performance of each backbone.

\begin{figure}[ht!]
     \centering
     \begin{subfigure}[b]{\linewidth}
         \centering
         \includegraphics[width=\linewidth]{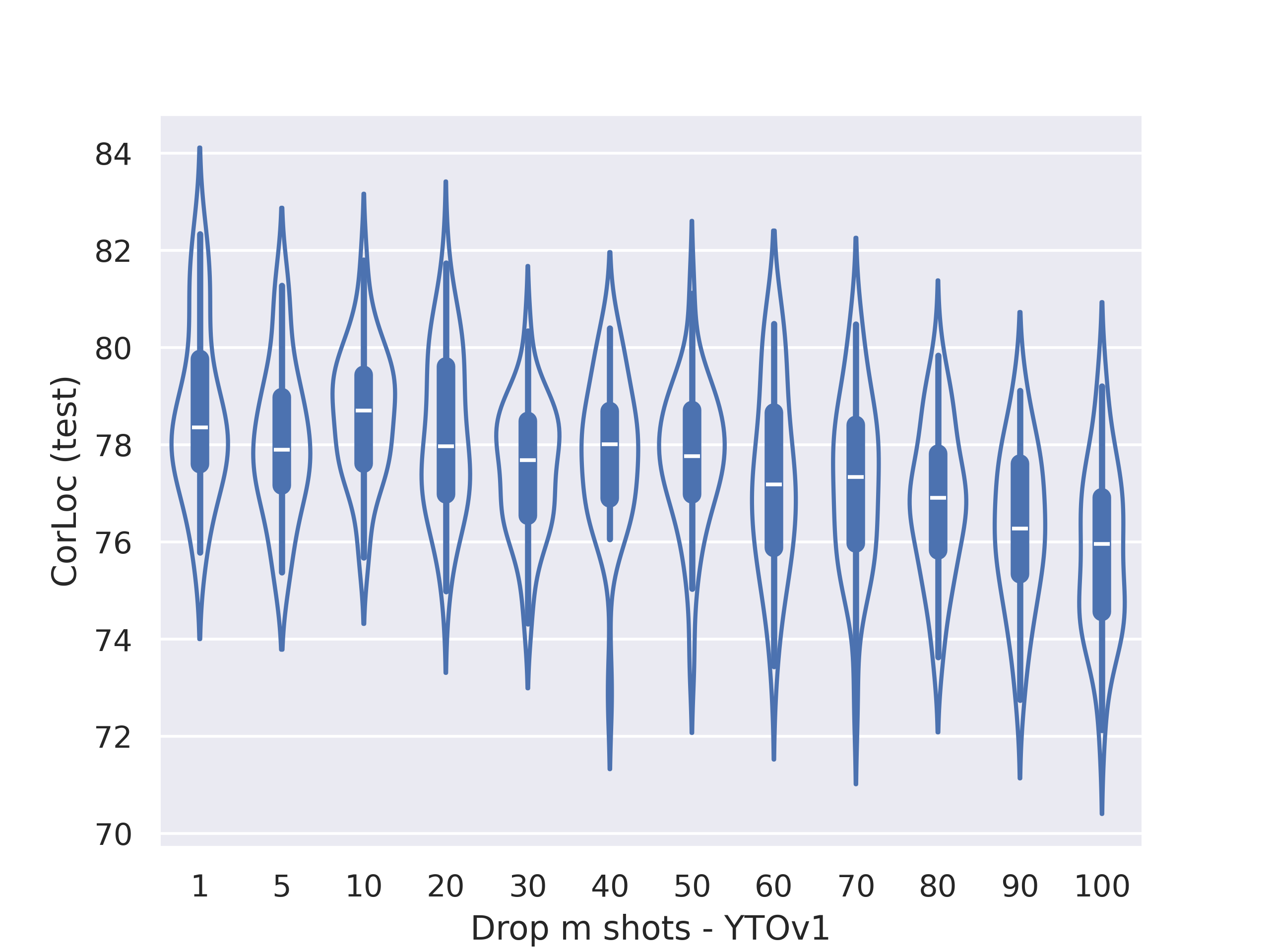}
         \caption{Performance stability: \ytovone}
         \label{fig:stability-yv1}
     \end{subfigure}
    \\
     \begin{subfigure}[b]{\linewidth}
         \centering
         \includegraphics[width=\linewidth]{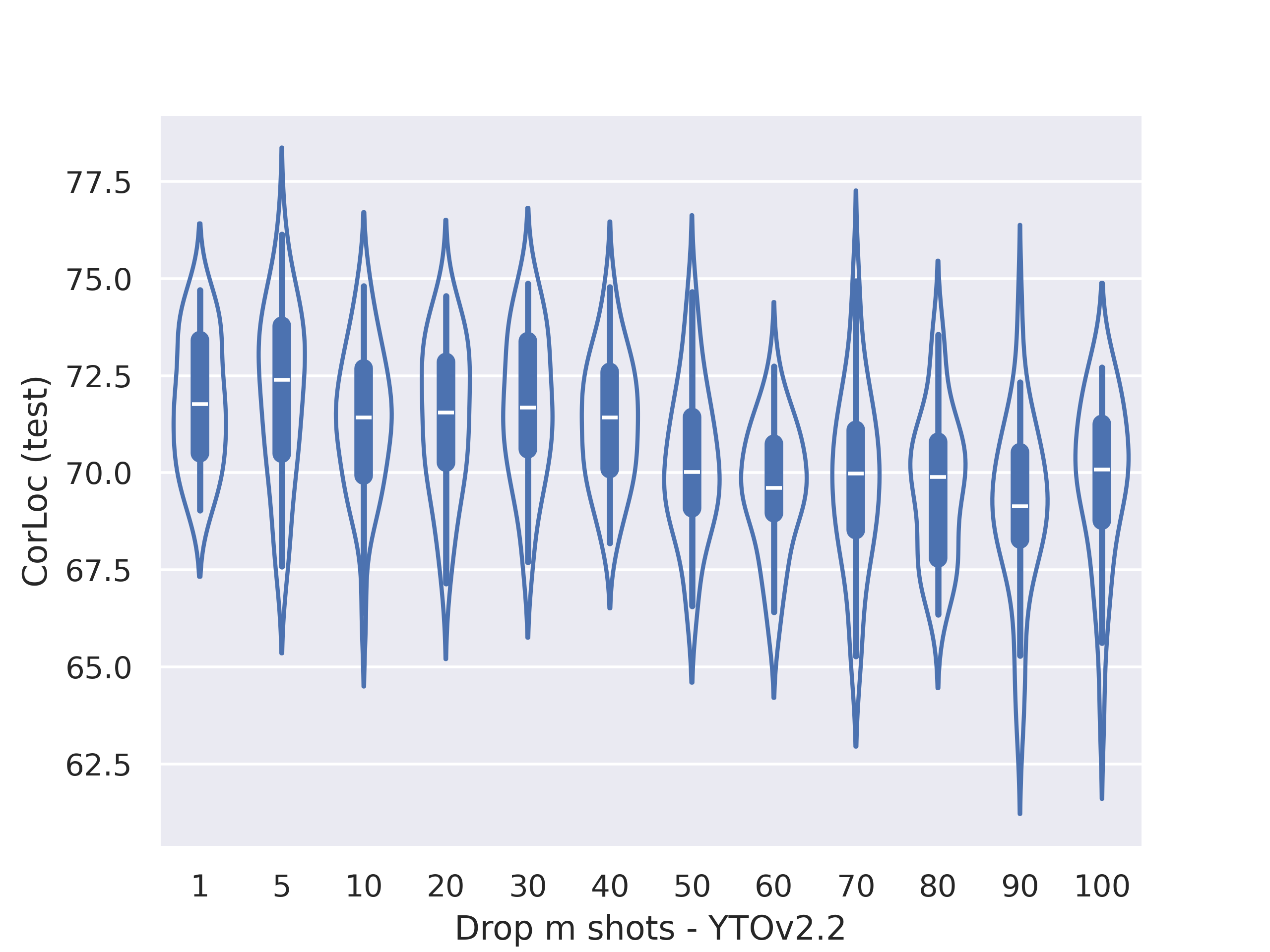}
         \caption{Performance stability: \ytovtwodtwo}
         \label{fig:stability-yv22}
     \end{subfigure}
        \caption{
        Stability of \colocam localization performance (\corloc) on  \ytovone and \ytovtwodtwo test set. Results are obtained when randomizing the effective training set by removing $m$ random shots from the train set. This process is repeated 30 times.
        }
        \label{fig:stability}
\end{figure}

\noindent\textbf{g) Performance stability (Fig.\ref{fig:ablation-lambda-c-const-vs-adap}).}
To further study the performance stability of our method, we conducted a new experiment by randomizing the effective train set since there are only very few labeled videos for validation set to be changed. Therefore, we keep the validation and test set fix. We only change the effective train set by discarding random shots for the main train set used so far. More specifically, we randomly removed $m$ shots per-class from the effective train set. We set ${m \in \{1, 5, 10, 20, 30, 40, 50, 60, 70, 80, 90, 100\}}$. Note that each video has many shots (\ie, short duration clip). This leads to a new effective train set which is less favorable to our method since it has less data. However, the purpose of this experiment is to check stability of our method. Consequently, the obtained results are not directly comparable to our main result with full train set nor to previous works. This new experiment consists of repeating the process of discarding $m$ random shots 30 times over both datasets \ytovone and \ytovtwodtwo using a fixed hyper-parameters set making the train set videos the only variable. Figure.\ref{fig:stability} shows the obtained results. We observe that in general, the results are fairly stable as they exhibit low standard deviation (Tab.\ref{tab:stability}). This last one varies from ${1.30}$ to ${1.72}$ over \ytovone, and between ${1.6}$ to ${2.26}$ over \ytovtwodtwo. Note that even when we drop 100 shots per class, our method still yields descent performance. However, as expected, reducing the size of effective train set leads to a drop in performance in general.

{
\begin{table}[ht!]
\centering
\resizebox{\linewidth}{!}{%
\centering
\begin{tabular}{lc|cc}
\hline
$m$ / Dataset && \ytovone & \ytovtwodtwo   \\
\hline \hline 
1   && $78.68 \pm 1.72$ & $71.84 \pm 1.66$ \\
5   && $78.07 \pm 1.55$ & $72.01 \pm 2.17$ \\
10   && $78.65 \pm 1.33$ & $71.35 \pm 1.84$ \\
20   && $78.26 \pm 1.63$ & $71.43 \pm 1.90$ \\ 
30   && $77.48 \pm 1.30$ & $71.69 \pm 1.89$ \\
40   && $77.84 \pm 1.52$ & $71.31 \pm 1.63$ \\
50   && $77.73 \pm 1.45$ & $70.30 \pm 1.92$ \\
60   && $77.35 \pm 1.86$ & $69.59 \pm 1.60$ \\
70   && $77.22 \pm 1.72$ & $69.81 \pm 2.26$ \\
80   && $76.85 \pm 1.50$ & $69.63 \pm 1.84$ \\
90   && $76.28 \pm 1.57$ & $69.02 \pm 2.19$ \\
100  && $75.77 \pm 1.68$ & $69.81 \pm 2.11$ \\
\hline 
\end{tabular}
}
\caption{Stability of \colocam localization performance (\corloc) on a test set obtained when randomizing the effective training set by removing $m$ random shots from the train set. This process is repeated 30 times. We report the mean ${\pm}$ standard deviation of \corloc accuracy for each $m$.}
\label{tab:stability}
\vspace{-1em}
\end{table}
}

\subsection{Analysis of CLIP-ES~\cite{lin2023} Localization Performance}

In this section, the performance of CLIP-ES~\cite{lin2023} is analyzed under different settings for the WSVOL task. CLIP-ES is a prompt-based method trained on a large corpus of text and image datasets. This model takes the image as input, in addition to the class label (text) to localize the ROI of the concerned class. 
CLIP-ES leverages two major supporting elements giving it more advantage compared to other methods:
1) Background prior list: It contains a list of objects that are potential background that are used to suppress and clean the initial CAM. Example of background classes\footnote{We have used the default background list provided in CLIP-ES code: $\{$'ground', 'land', 'grass', 'tree', 'building', 'wall', 'sky', 'lake', 'water', 'river', 'sea', 'railway', 'railroad', 'keyboard', 'helmet', 'cloud', 'house', 'mountain', 'ocean', 'road', 'rock', 'street', 'valley', 'bridge', 'sign'$\}$.}: 'ground', 'grass'. 
2) Refine CAM: After suppressing the background, CAM is then post-processed to improve its localization performance.

Note that all other methods do not have access to both options, making them in a disadvantage. For a fair comparison, we report the CLIP-ES performance using only the initial CAM. From Tab.\ref{tab:clip-es}, it is clear that a major improvement in localization comes from the background list and CAM refinement. Initial CAM has a performance of ${65.1\%, 60.2\%}$ over \ytovone, and \ytovtwodtwo, respectively. Using these two post-processing techniques allows for a jump in localization performance to ${88.2\%, 84.8\%}$ on both datasets. 

Applying CLIP-ES in WSVOL/WSOL tasks has a major issue in real-world applications regarding the class label since it is required as \emph{input}. Up to now, we used the true class provided as annotation (\labelgt). Since this model does not perform a classification task, we experiment with a scenario where a pre-trained classifier is used to predict the class in the image. Then, the predicted class is used as input to CLIP-ES. We report the performance in Tab.\ref{tab:clip-es} with the image label named \labelpr. The results show a clear drop in localization performance which is explained by the errors in the predicted classes. Note that we trained a classifier over each dataset using the ResNet50 backbone. We obtained ${72.5\%}$ of classification accuracy over \ytovone test set, and ${68.5\%}$ over \ytovtwodtwo test set. We extend this scenario to a poor classifier (uniform) by marginalizing localization performance over all possible classes (\labelavg). This is done by setting the input class as constant and averaging the localization performance over all the classes.

{
\begin{table}[ht!]
\centering
\resizebox{\linewidth}{!}{%
\centering
\begin{tabular}{|ccccc|cc|}
\hline
\multicolumn{4}{|c}{Configurations} && \multicolumn{2}{c|}{Datasets} \\
\hline
Initial  & List prior   & Image & Refined  & &  &  \\
CAM  & background &  label  & CAM & &  \ytovone & \ytovtwodtwo  \\
\hline 
\cmark & \xmark                & \labelgt & \xmark           & &  65.1     & 60.2  \\
\cmark & \cmark                & \labelgt & \xmark           & &  76.6     & 72.6  \\
\cmark & \cmark                & \labelgt & \cmark           & &  88.2     & 84.8  \\
\hline
\cmark & \xmark                & \labelpr & \xmark           & &  56.7     & 50.7  \\
\cmark & \cmark                & \labelpr & \xmark           & &  70.9     & 64.9  \\
\cmark & \cmark                & \labelpr & \cmark           & &  83.9     & 78.9  \\
\hline
\cmark & \xmark                & \labelavg & \xmark           & &  22.0     & 21.7  \\
\cmark & \cmark                & \labelavg & \xmark           & &  51.9     & 44.6  \\
\cmark & \cmark                & \labelavg & \cmark           & &  68.7     & 63.7  \\
\hline
\end{tabular}
}
\caption{\corloc localization accuracy of CLIP-ES~\cite{lin2023} method with different configurations on \ytovone and \ytovtwodtwo test sets. Labels: \labelgt: ground truth class label. \labelpr: class label predicted by a trained classifier over the corresponding dataset. \labelavg: average \corloc localization over all labels over a single sample.}
\label{tab:clip-es}
\vspace{-1em}
\end{table}
}

\begin{figure*}
     \centering
     \begin{subfigure}[b]{0.24\linewidth}
         \centering
         \includegraphics[width=\linewidth]{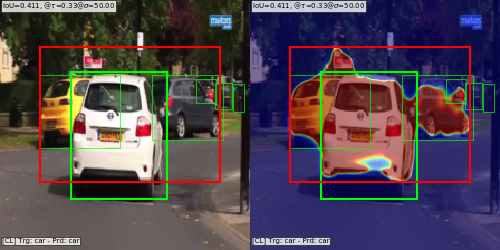}
     \end{subfigure}
     \begin{subfigure}[b]{0.24\linewidth}
         \centering
         \includegraphics[width=\linewidth]{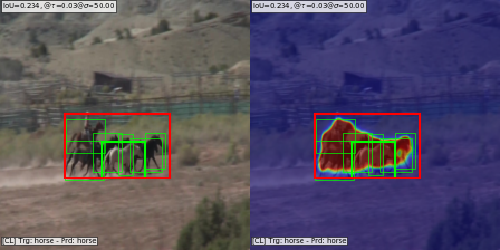}
     \end{subfigure}
     \begin{subfigure}[b]{0.24\linewidth}
         \centering
         \includegraphics[width=\linewidth]{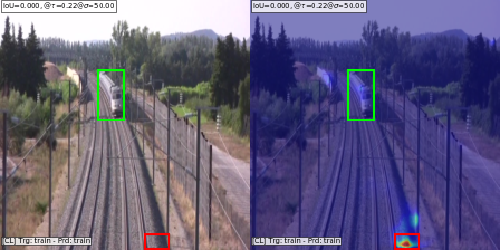}
     \end{subfigure}
     \begin{subfigure}[b]{0.24\linewidth}
         \centering
         \includegraphics[width=\linewidth]{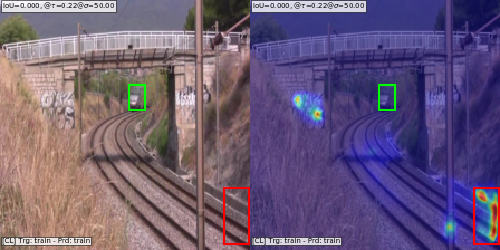}
     \end{subfigure}
        \caption{Typical challenges of our method. \emph{Column 1-2}: Dense and overlapped instances lead to fused localization of instances. \emph{Column 3-4}: Object mis-localization. Classifier CAM does not activate over the right object. Our trained decoder using these CAMs is unable to correct this large error. Bounding boxes: ground truth (green), prediction (red).}
        \label{fig:failure}
\end{figure*}

\begin{figure}
     \centering
     \begin{subfigure}[b]{0.44\textwidth}
         \centering
         \includegraphics[width=\textwidth]{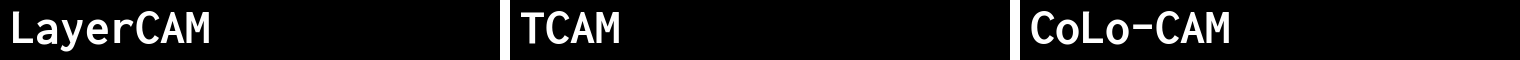}
     \end{subfigure}
     \\
     \begin{subfigure}[b]{0.44\textwidth}
         \centering
         \includegraphics[width=\textwidth]{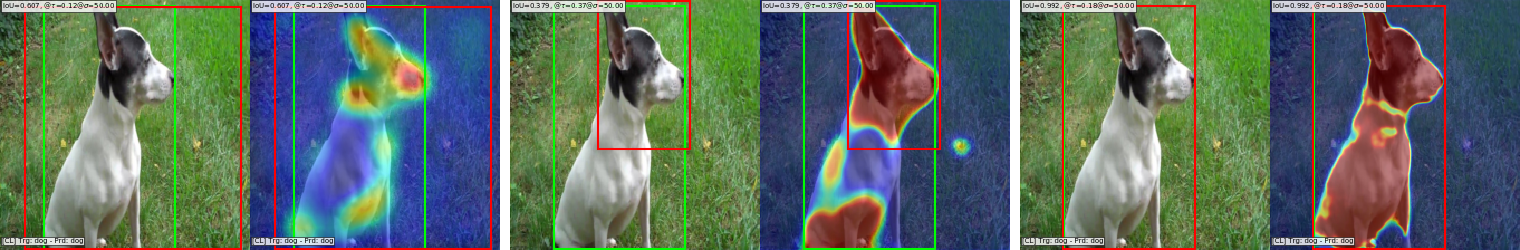}
     \end{subfigure}
     \\
     \begin{subfigure}[b]{0.44\textwidth}
         \centering
         \includegraphics[width=\textwidth]{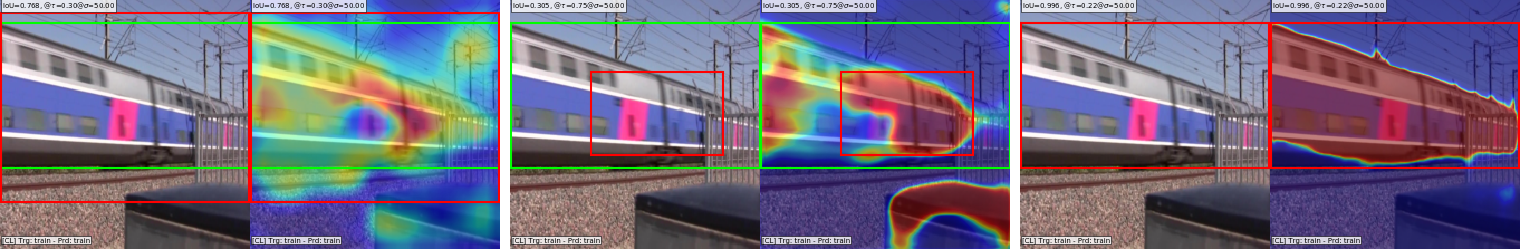}
     \end{subfigure}
     \\
     \begin{subfigure}[b]{0.44\textwidth}
         \centering
         \includegraphics[width=\textwidth]{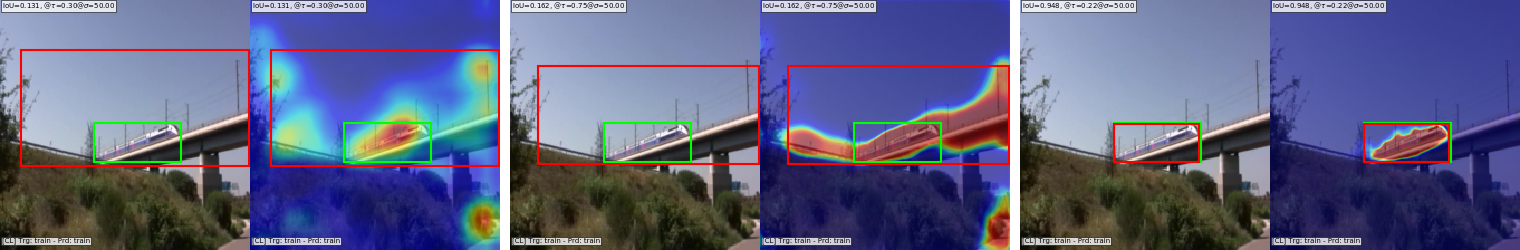}
     \end{subfigure}
     \\
     \begin{subfigure}[b]{0.44\textwidth}
         \centering
         \includegraphics[width=\textwidth]{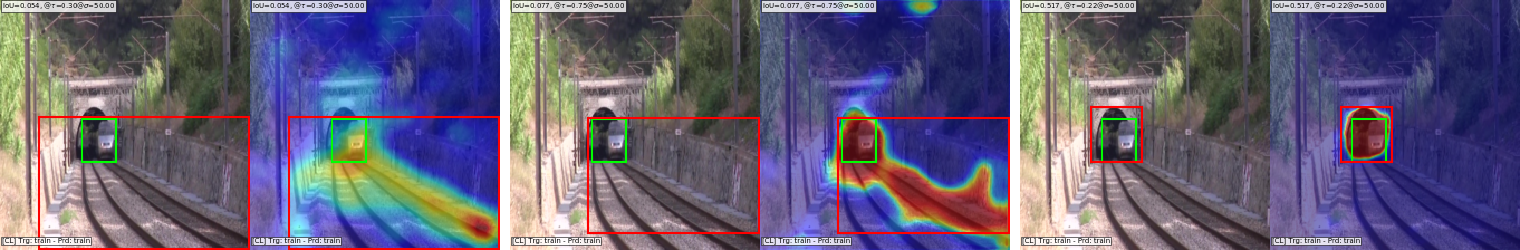}
     \end{subfigure}
      \\
     \begin{subfigure}[b]{0.44\textwidth}
         \centering
         \includegraphics[width=\textwidth]{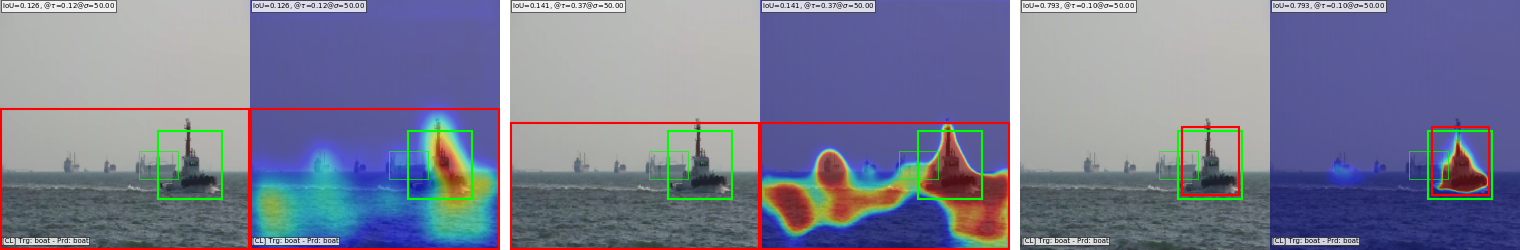}
     \end{subfigure}
     \\
     \begin{subfigure}[b]{0.44\textwidth}
         \centering
         \includegraphics[width=\textwidth]{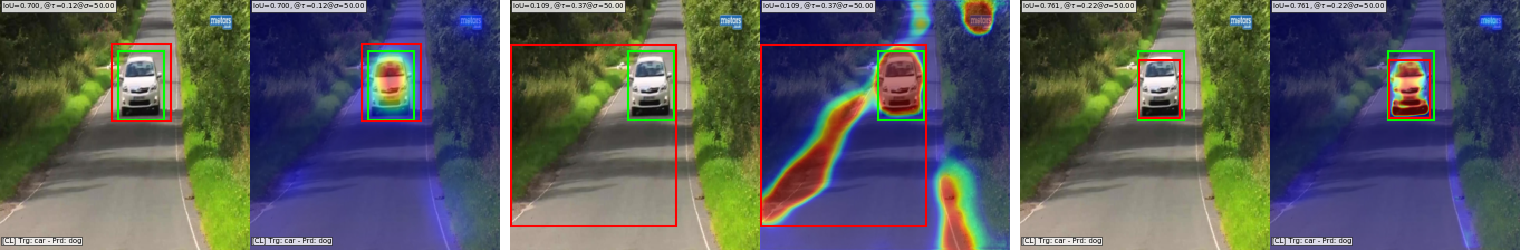}
     \end{subfigure}
     \\
     \begin{subfigure}[b]{0.44\textwidth}
         \centering
         \includegraphics[width=\textwidth]{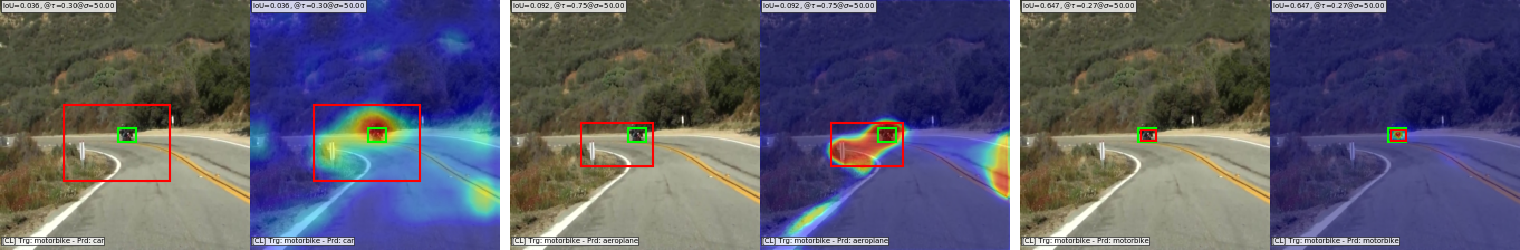}
     \end{subfigure}
     \\
     \begin{subfigure}[b]{0.44\textwidth}
         \centering
         \includegraphics[width=\textwidth]{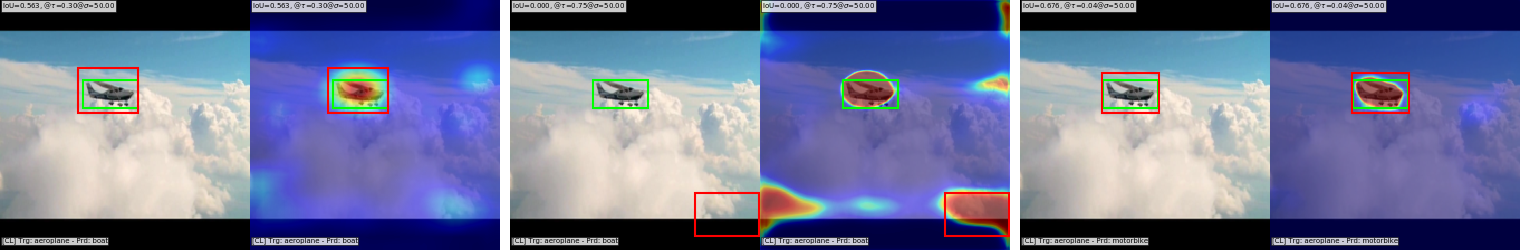}
     \end{subfigure}
     \\
     \begin{subfigure}[b]{0.44\textwidth}
         \centering
         \includegraphics[width=\textwidth]{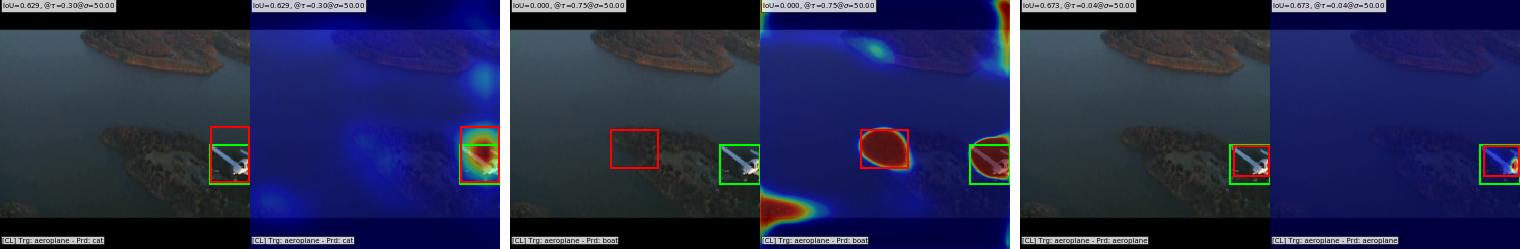}
     \end{subfigure}
        \caption{Localization examples of test sets frames. 
        \emph{Bounding boxes}: ground truth (green), prediction (red). The second column of each method is the predicted CAM over image.}
        \label{fig:visu-pred}
\end{figure}

\subsection{Visual Results}

Compared to TCAM~\cite{tcamsbelharbi2023} and LayerCAM~\cite{JiangZHCW21layercam}, we observe that our CAMs became more discriminative (Fig.\ref{fig:visu-pred}). This translates into several advantageous properties. Our CAMs showed sharp, more complete, and less noisy activations localized more precisely around objects. In addition, localizing small objects such as 'Bike', and large objects such as 'Train' becomes more accurate. CAM activations often suffer from co-occurrence issues~\cite {tcamsbelharbi2023,rony2023deep} where consistently appearing contextual objects (background) are confused with main foreground objects. Our CAMs showed more robustness to this issue. This can be seen in the case of road vs. 'Car'/'Bike', or water vs. 'Boat'.

Despite this improvement, our method still faces two main challenges (Fig.\ref{fig:failure}). The first is the case of samples with dense and overlapped instances (top Fig.\ref{fig:failure}). This often causes activations to spill across instances to form a large object hindering localization performance. The second issue concerns object mis-localization (bottom Fig.\ref{fig:failure}). Since our localization is mainly stimulated by discriminative cues from CAMs, it is still challenging to detect when a classifier points to the wrong object. This often leads to large errors that are difficult to correct in downstream use. Future work may consider leveraging the classifier response over ROIs to ensure their quality. Additionally, object movement information can be combined with CAM cues.

Demonstrative videos provided in the code highlight additional challenges that are not visible in still images, in particular:  

\noindent\emph{1) Empty frames}. In unconstrained videos, the labeled object in the video is not necessarily present in the entire video. Therefore, some frames may not contain the object. Despite this, classifier CAMs still show class activations on these frames. This issue is the result of the weak global annotation of videos, and the assumption that all frames inherit that same label. It highlights the importance of detecting such frames and treating them accordingly, starting with classifier training. Discarding these frames will help the classifier better discern objects and reduce noisy CAMs. Without any additional supervision, this adds more difficulty to WSVOL task. One possible solution to this issue is to leverage the per-class probabilities of the classifier to assess whether an object is present or not in a frame. Such likelihood can be exploited to discard frames or weight loss terms over them.

\noindent\emph{2) Temporal consistency}. Although our method brought a quantitative and qualitative improvement, there are some failure cases. In some instances, we observed inconsistencies between consecutive frames. This error is characterized by a large shift in localization where the bounding box (and the CAM activation), moves abruptly from one location at a frame to completely different location at the next frame. This often happens when objects become partially or fully occluded, the appearance of other instances, the scene becomes complex, or zooming in the image makes the object too large. Such setups lead to a \emph{sudden shift} of the CAM's focus, leading to an abrupt change of localization which is undesirable. This drawback is mainly inherited from the \emph{inference} procedure performed at a single frame which does not account for spatiotemporal information allowing such errors. Future works may consider improving the inference procedure by leveraging spatiotemporal information at the expense of inference time.

\section{Conclusion}
\label{sec:conclusion}

This paper proposes a new CAM-based approach for WSVOL by introducing co-localization into a CAM-based method. Using pixel-color cue, our proposed \colocam method constrains CAM responses to be similar over visually similar pixels across all frames, achieving col-localization.
Our total training loss comprises per-frame and multi-frame terms optimized simultaneously via standard Stochastic Gradient Descent (SGD).
Empirical experiments showed that our produced CAMs become more discriminative. This translates into several advantageous properties: sharp, more complete, and less noisy activations localized more precisely around objects. Moreover, localizing small/large objects becomes more accurate. Additionally, our method is more robust to training with long time-dependency and leads to new state-of-the-art localization performance.

Despite its improvements, our method still holds some limitations. Our pixel pseudo-labels are less accurate since they come from a pretrained classifier. In particular, a major issue is the set of frames without object of interest in a video. In WSVOL, it is typically assumed that the video label transfers to each frame. However, in unconstrained videos, not all frames hold the labeled object. Therefore, the label is mistakenly transferred into empty frames. This furthermore allows the classifier to suggest relevant ROIs in such frames leading to more errors in training. One possible way to alleviate this issue is to leverage the per-class probabilities of the classifier to assess whether an object is present or not in a frame. Such likelihood can be exploited to discard frames or weight loss terms over them.

Another issue is related to inference that is performed frame by by frame without leveraging temporal information in a video. The negative impact of this can be observed over a sequence of predictions which are characterized by localization inconsistencies. This is illustrated by large shift in localization where the bounding box (and the CAM activation), moves abruptly from one location at a frame to completely different location at the next frame. Prediction without accounting for temporal dependencies can easily lead to such results. Future works may consider improving the inference procedure by leveraging spatiotemporal information at the expense of inference time.

We note that our method can use either CNN- or transformer-based architecture. The only requirement is that the architecture is able to perform both tasks: classification and localization. 
A vision transformer-based (ViT) model~\cite{dosovitskiy21} can be employed as an encoder. A standard classification head can be used to perform classification task. A new localization head is required to replace our decoder to perform localization task. An architecture similar to the one  presented in~\cite{Murtaza2023dips} can be used in our work which is expected to yield better performance compared to CNN-based model as reported in their work. We leave this aspect for future work.

\section*{Acknowledgment}
This research was supported in part by the Canadian Institutes of Health Research, the Natural Sciences and Engineering Research Council of Canada, and the Digital Research Alliance of Canada (alliancecan.ca).


\appendices

%
%
%
%

\begin{center}
\IncMargin{0.04in}
\begin{algorithm*}
    \SetKwInOut{Input}{Input}
    \Input{
    ${\bm{X}_t}$: Input frame at time $t$ with its image label $y$, and its corresponding upscaled CAM ${\bm{C}_t}$ from a pre-trained classifier. 
    \\
    ${f}$: Localization decoder with parameters  ${\bm{\theta}}$.
    \\
    ${\texttt{max\_epochs}: \textrm{Maximum number of training epochs}}$.
    }
    \vspace{0.1in}
    Define foreground region ${\mathbb{C}^+_t}$ by thresholding the CAM ${\bm{C}_t}$ using Ostu method. \\
    \vspace{0.025in}
    Define foreground region ${\mathbb{C}^-_t}$ as the complement of the foreground: ${\mathbb{C}^-_t = \Omega - \mathbb{C}^+_t}$ where ${\Omega}$ is the entire image locations. \\
    \vspace{0.025in}
    Define ${\mathcal{M}(\mathbb{C}^+_t)}$ as a multinomial sampling distribution over the foreground region. It uses the CAM normalized scores as sampling probabilities.
    \\
    \vspace{0.025in}
    Define ${\mathcal{U}(\mathbb{C}^-_t)}$ as a  uniform distribution over the background region.\\
    \vspace{0.1in}
    \For{${i \in 1 \cdots \texttt{max\_epochs}}$} 
    {
        \colorbox{mybluelight}{
        Sample a pixel location from foreground using ${\mathcal{M}(\mathbb{C}^+_t)}$, and a background pixel
        } 
        \colorbox{mybluelight}{
        using ${\mathcal{U}(\mathbb{C}^-_t)}$.
        }
        \\
        \vspace{0.03in}
        Construct a pseudo-label mask ${Y_t}$, where ${Y_t(p) \in \{0, 1\}^2}$, where $0$ is for background pixels, and $1$ is for foreground pixels. Pixels that have not been selected in this iteration are set to be unknown.\\
        \vspace{0.03in}
        Perform an SGD step over total training loss using Eq.\ref{eq:totalloss}. The mask ${Y_t}$ is used only with partial cross-entropy loss defined in Eq.\ref{eq:pl}.
    }
    \vspace{0.1in}
    \caption{
    Random pixel samples selection during training illustrated over a single image. \colorbox{mybluelight}{Blue background} indicates that randomness changes for each frame, and each epoch.
    }
    \label{alg:pixels-sampling}
\end{algorithm*}
\DecMargin{0.04in}
\end{center}

\section{Proposed Approach}
\label{sec:method-supp-mat}

\subsection{Detailed Illustrations for Training and Inference} 
\label{subsec:app-train-inf}

\begin{figure*}[ht!]
\centering
  \centering
  \includegraphics[width=\linewidth]{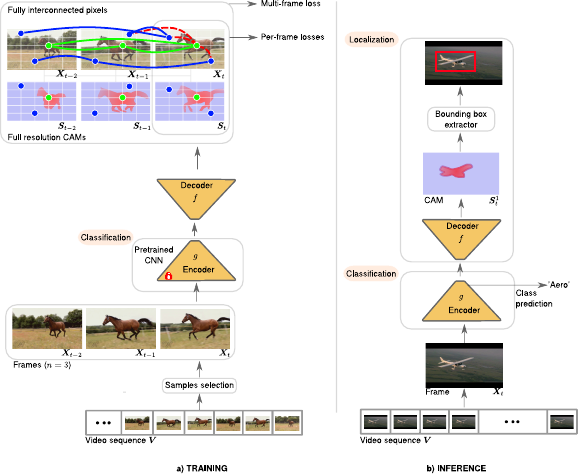}
  \caption[Caption]{Training and inference with our proposed CoLo-CAM method. The single-frame and multi-frame training loss terms are illustrated with $n=3$ frames. \emph{Left: training phase}: It combines the per-frame terms composed of pseudo-labels loss (PL), absolute size constraint (ASC), and CRF loss, as well as the multi-frame term.
  \emph{Right: inference phase}: It operates at a single frame with no temporal dependency, and predicts bounding box localization and classification.
  Notation and per-frame and multi-frame terms are described in Section \ref{sec:method}.}
  \label{fig:proposal-details}
\end{figure*}

Fig.\ref{fig:proposal-details} presents our detailed model for training and inference. During the training, we consider ${n=3}$ input frames. We use three per-frame terms that aim at improving localization at each frame, and without considering any spatiotemporal information. Using the classifier CAM ${\bm{C}_t}$, we sample pseudo-labels which are used at the pseudo-label term (PL). Additionally, we apply a CRF, and a size constraint (ASC). In parallel, we apply a multi-frame term ver all the frames allowing explicit communication between them. This is achieved by constraining the output CAMs of the frames to activate similarly over pixels with similar color.

At inference time, we simply consider single frames, \ie, no spatiotemporal information. A bounding box is estimated from the foreground CAM, at the decoder level. In addition, frame classification scores are obtained from the encoder.

\subsection{Proposed Approach: Learning with Pixel Pseudo-Labels}
\label{subsec:app-pseudp-label}

In this section, we provide additional details on sampling PLs through the classifier's CAM ${\bm{C}_t}$. These sampled pseudo-labels are used to train the decoder $f$, in combination with other terms. Through the discriminative property of the pseudo-labels, they initiate and stimulate object localization. In CAM-based methods for WSOL, it is common to assume that CAMs' strong activations are likely to point to the foreground, whereas low activations hint at background regions~\cite{zhou2016learning}.

In the following, we denote the foreground region as ${\mathbb{C}^+_t}$ which is computed by the operation ${\mathcal{O}^+}$. For simplicity and without adding new hyper-parameters, ${\mathbb{C}^+_t}$ is defined as the set of pixels with CAM activation ${\bm{C}_t}$ greater than a threshold estimated automatically by Otsu method over ${\bm{C}_t}$. The background region, \ie, ${\mathbb{C}^-_t}$, is the remaining set of pixels. Both regions are defined as follows,
\begin{equation}
    \label{eq:sets}
    \mathbb{C}^+_t = \mathcal{O}^+(\bm{C}_t), \quad \mathbb{C}^-_t = \Omega - \mathbb{C}^+_t \;, 
\end{equation}
where ${\Omega}$ is a discrete image domain. 
Given the weak supervision, these regions are uncertain. For instance, the ${\mathbb{C}^+_t}$ region may hold only a small part of the foreground, along with the background parts. Consequently, the background region ${\mathbb{C}^-_t}$ may still have foreground parts. This uncertainty in partitioning the image into foreground and background makes it unreliable for localization to directly fit the model to entire regions, as it is commonly done~\cite{KolesnikovL16}. Instead, we pursue a stochastic sampling strategy in both regions to avoid overfitting and to provide the model enough time for the emergence of consistent regions. This has shown to be a better strategy~\cite{belharbi2022fcam,negevsbelharbi2022}. For each frame, and an SGD step, we sample a pixel from ${\mathbb{C}^+_t}$ as the foreground, and a pixel from ${\mathbb{C}^-_t}$ as background to be pseudo-labels. We encode their location according to:
\begin{equation}
    \label{eq:sset}
    \Omega^{\prime}_t = \mathcal{M}(\mathbb{C}^+_t)\; \cup \; \mathcal{U}(\mathbb{C}^-_t) \;, 
\end{equation}
where ${\mathcal{M}(\mathbb{C}^+_t)}$ is a multinomial sampling distribution over the foreground, and ${\mathcal{U}(\mathbb{C}^-_t)}$ is a uniform distribution over the background. This choice of distributions is based on the assumption that the foreground object is often concentrated in one place, whereas the background region is spread across the image. The partially pseudo-labeled mask for the sample ${\bm{X}_t}$ is referred to as ${Y_t}$, where ${Y_t(p) \in \{0, 1\}^2}$ is the binary label of the location ${p}$ with value ${0}$ for the background and ${1}$ for the foreground. Undefined regions are labeled as unknown.
The pseudo-annotation mask ${Y_t}$ is then leveraged to train the decoder using partial cross-entropy. 
Such stochastic pseudo-labels are expected to stimulate the learning of the model and guide it to generalize and respond in the same way over regions with similar color/texture.
The sampling process of pixel pseudo-labels is summarized in Algorithm.\ref{alg:pixels-sampling}.

\section{Visual Results.} 
\label{sec:app-more-visuals}

\begin{figure}
     \centering
     \begin{subfigure}[b]{0.35\textwidth}
         \centering
         \includegraphics[width=\textwidth]{tag}
     \end{subfigure}
     \\
     \begin{subfigure}[b]{0.35\textwidth} 
 \centering 
 \includegraphics[width=\textwidth]{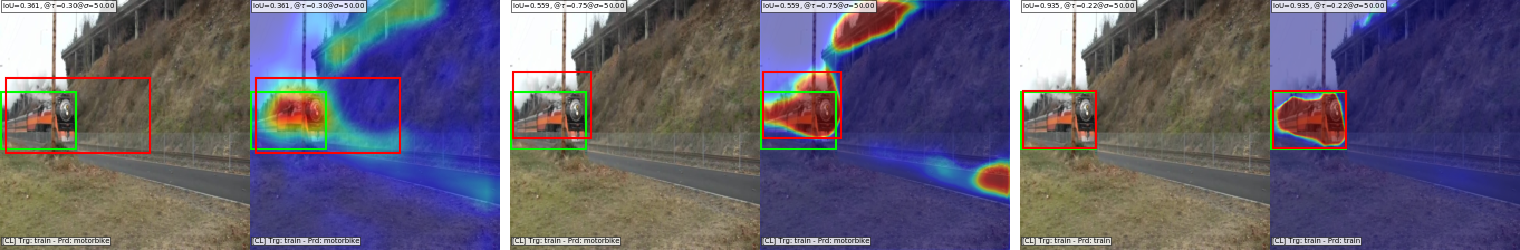} 
  \end{subfigure} 
 \\
\begin{subfigure}[b]{0.35\textwidth} 
 \centering 
 \includegraphics[width=\textwidth]{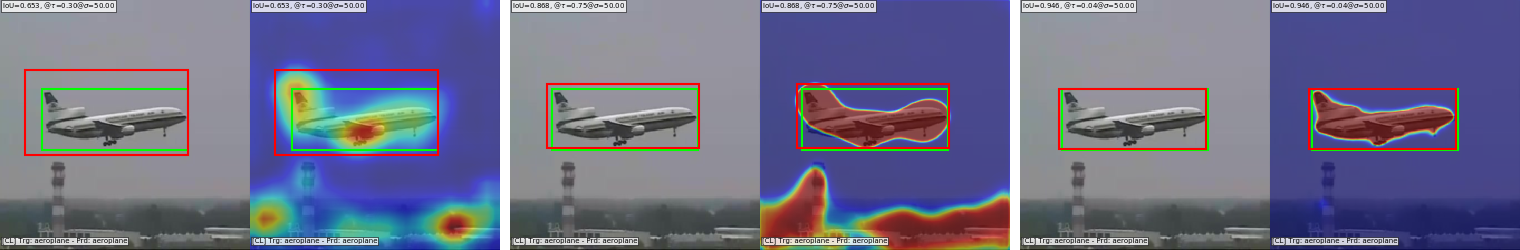} 
  \end{subfigure} 
 \\
\begin{subfigure}[b]{0.35\textwidth} 
 \centering 
 \includegraphics[width=\textwidth]{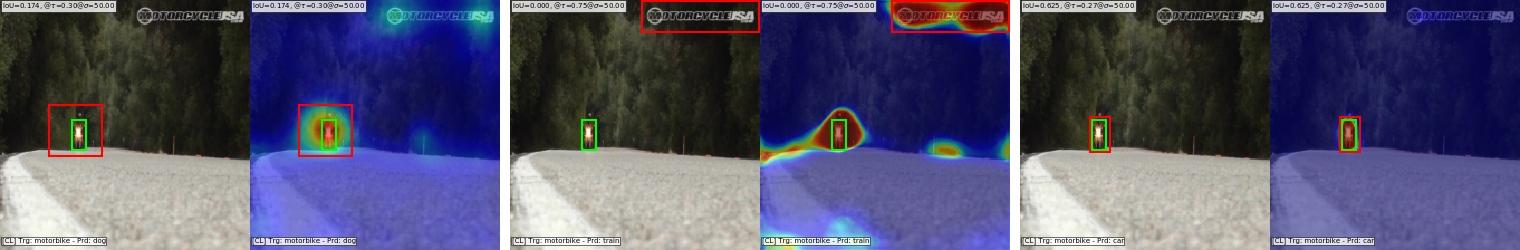} 
  \end{subfigure} 
 \\
\begin{subfigure}[b]{0.35\textwidth} 
 \centering 
 \includegraphics[width=\textwidth]{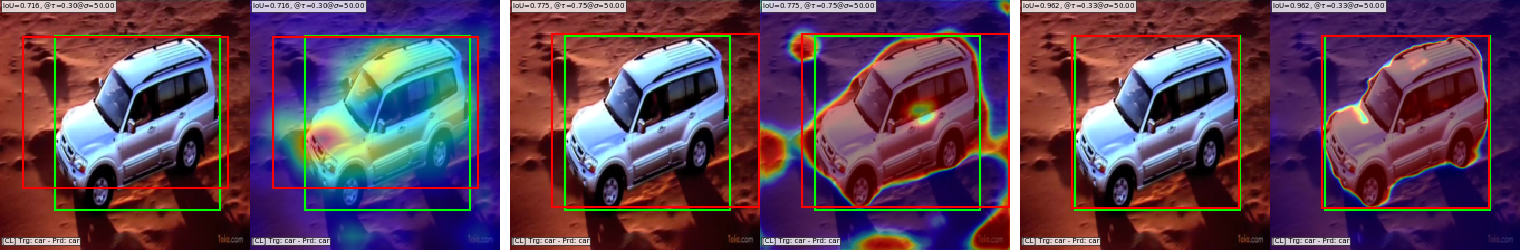} 
  \end{subfigure} 
 \\
\begin{subfigure}[b]{0.35\textwidth} 
 \centering 
 \includegraphics[width=\textwidth]{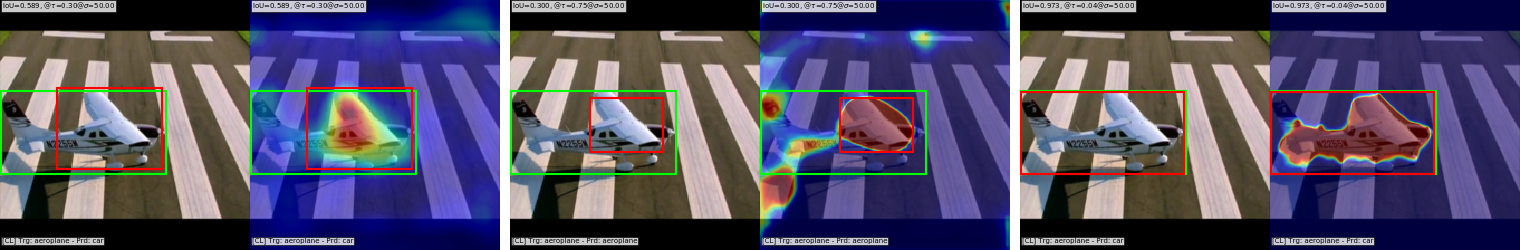} 
  \end{subfigure} 
 \\
\begin{subfigure}[b]{0.35\textwidth} 
 \centering 
 \includegraphics[width=\textwidth]{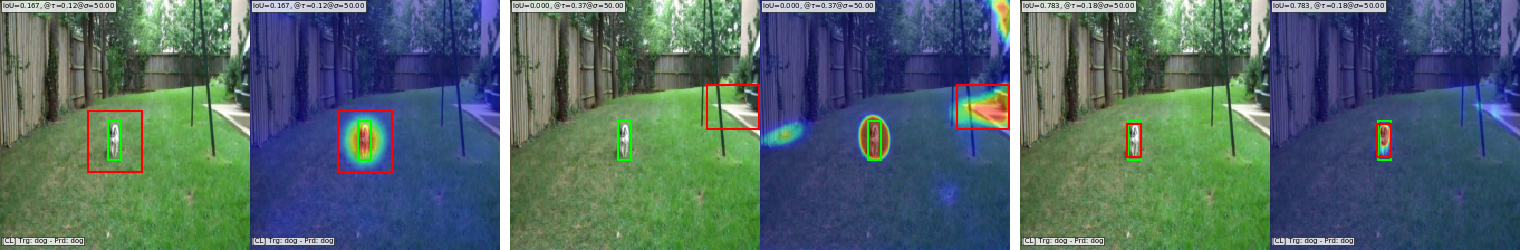} 
  \end{subfigure} 
 \\
\begin{subfigure}[b]{0.35\textwidth} 
 \centering 
 \includegraphics[width=\textwidth]{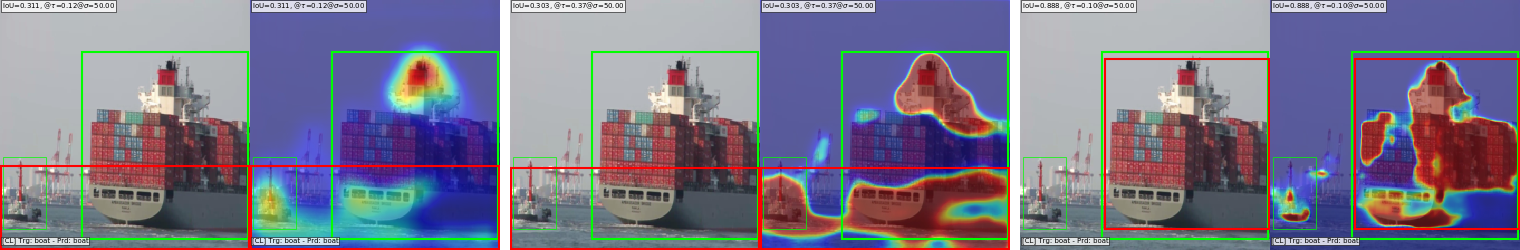} 
  \end{subfigure} 
 \\
\begin{subfigure}[b]{0.35\textwidth} 
 \centering 
 \includegraphics[width=\textwidth]{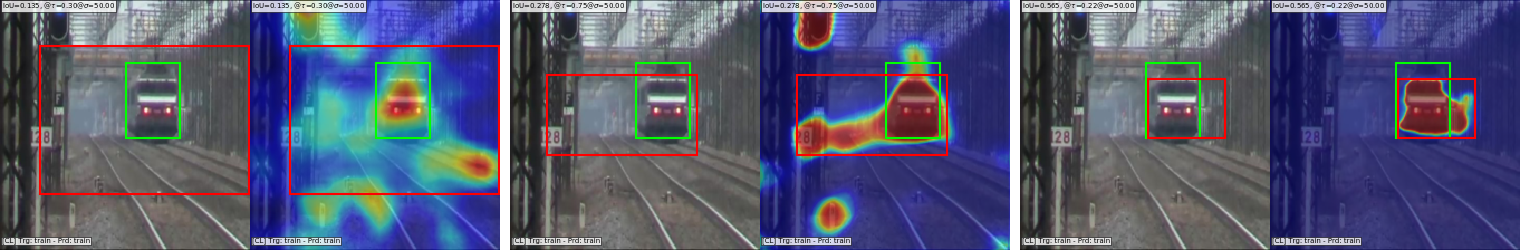} 
  \end{subfigure} 
 \\
\begin{subfigure}[b]{0.35\textwidth} 
 \centering 
 \includegraphics[width=\textwidth]{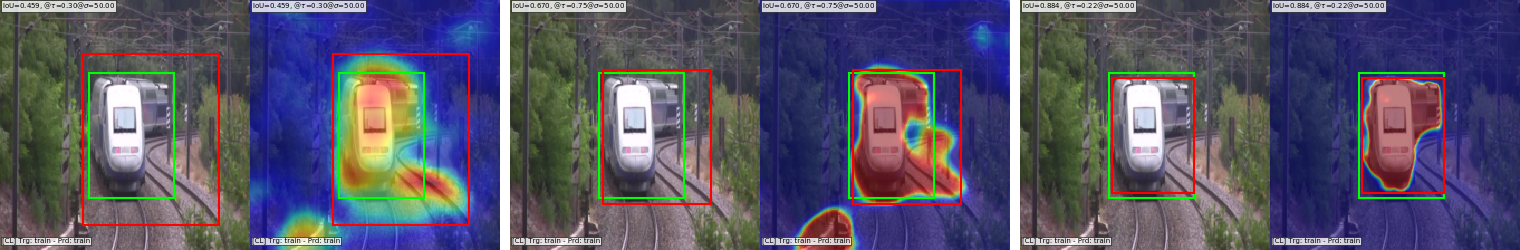} 
  \end{subfigure} 
 \\
\begin{subfigure}[b]{0.35\textwidth} 
 \centering 
 \includegraphics[width=\textwidth]{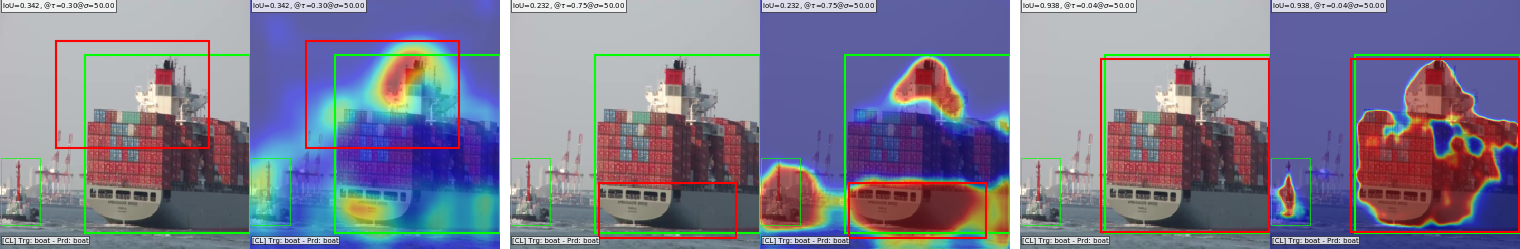} 
  \end{subfigure} 
 \\
\begin{subfigure}[b]{0.35\textwidth} 
 \centering 
 \includegraphics[width=\textwidth]{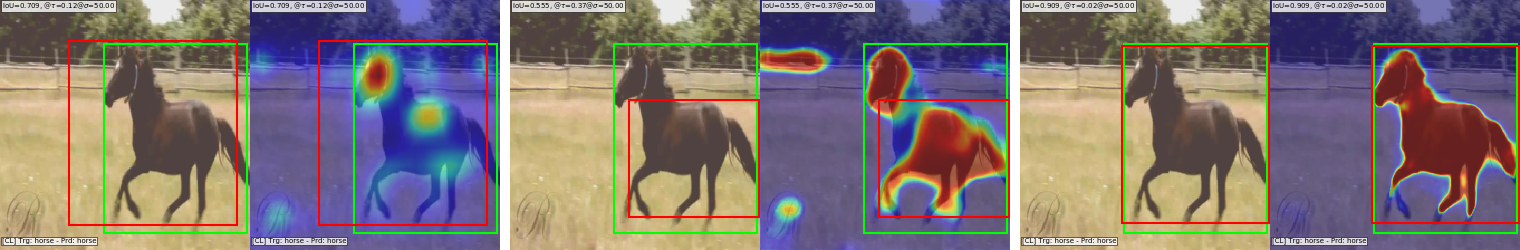} 
  \end{subfigure} 
 \\
\begin{subfigure}[b]{0.35\textwidth} 
 \centering 
 \includegraphics[width=\textwidth]{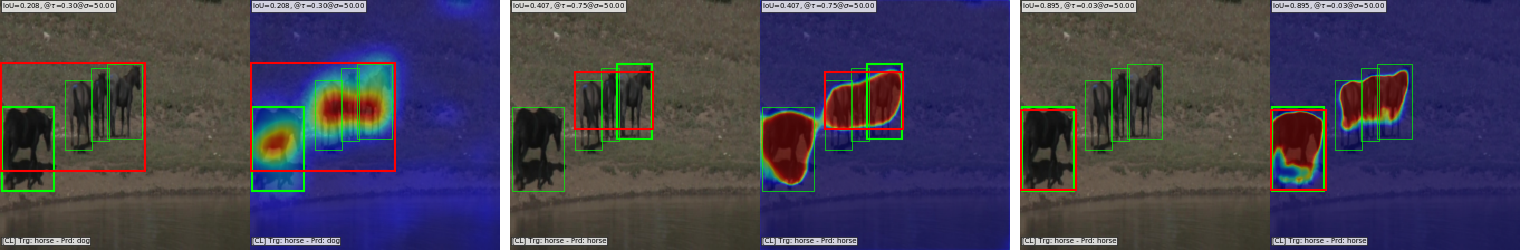} 
  \end{subfigure} 
 \\
\begin{subfigure}[b]{0.35\textwidth} 
 \centering 
 \includegraphics[width=\textwidth]{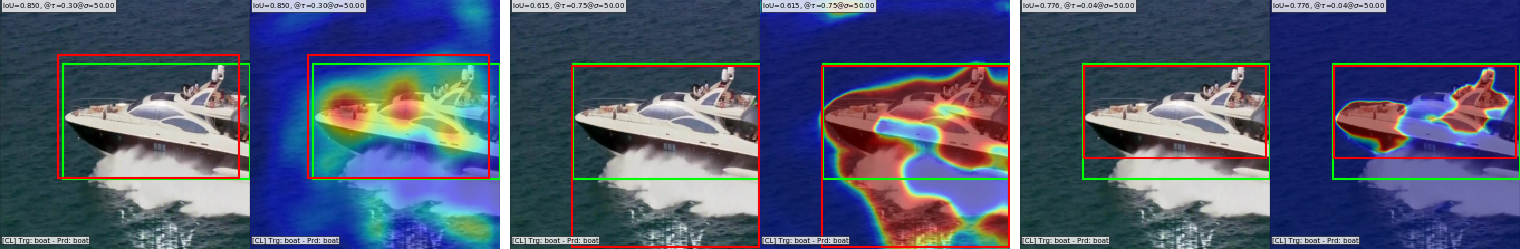} 
  \end{subfigure} 
 \\
\begin{subfigure}[b]{0.35\textwidth} 
 \centering 
 \includegraphics[width=\textwidth]{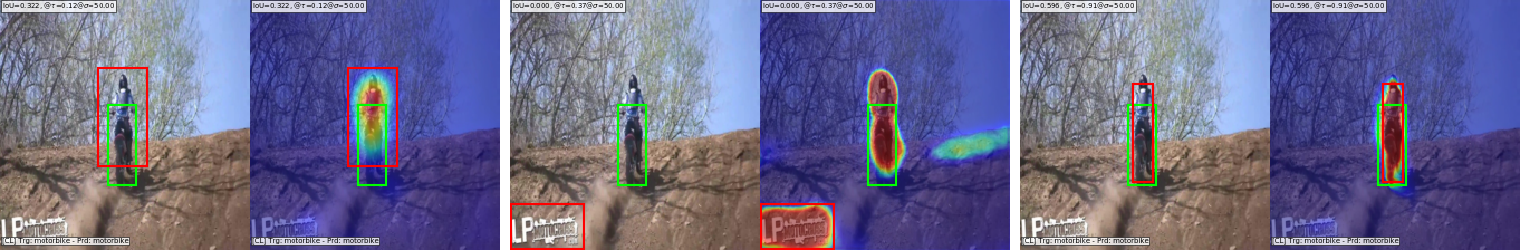} 
  \end{subfigure} 
 \\
\begin{subfigure}[b]{0.35\textwidth} 
 \centering 
 \includegraphics[width=\textwidth]{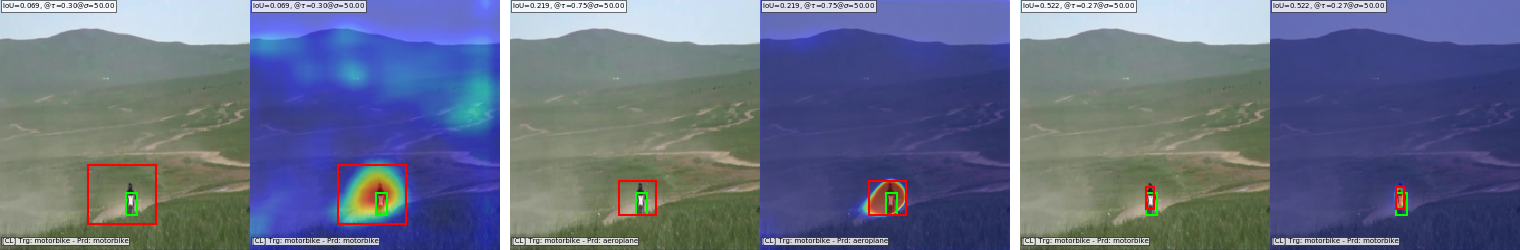} 
  \end{subfigure} 
 \\
\begin{subfigure}[b]{0.35\textwidth} 
 \centering 
 \includegraphics[width=\textwidth]{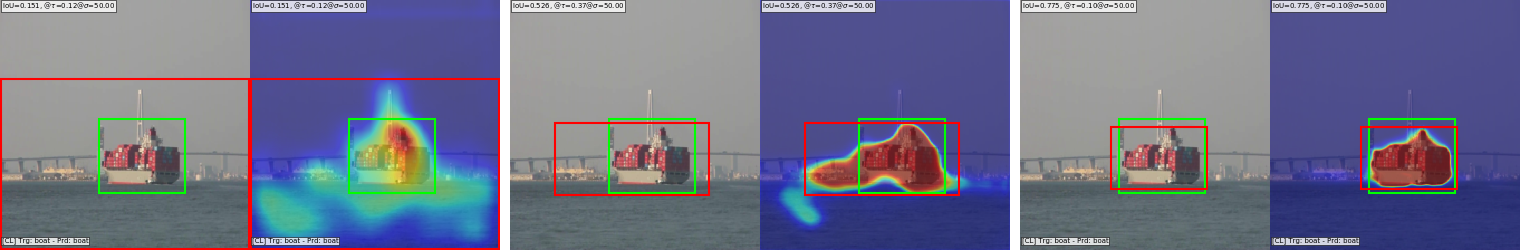} 
  \end{subfigure} 
 \\
\begin{subfigure}[b]{0.35\textwidth} 
 \centering 
 \includegraphics[width=\textwidth]{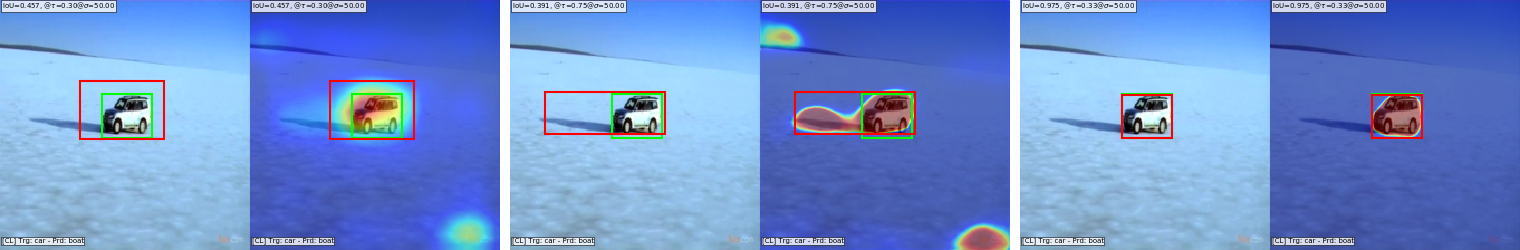} 
  \end{subfigure} 
 \\
\begin{subfigure}[b]{0.35\textwidth} 
 \centering 
 \includegraphics[width=\textwidth]{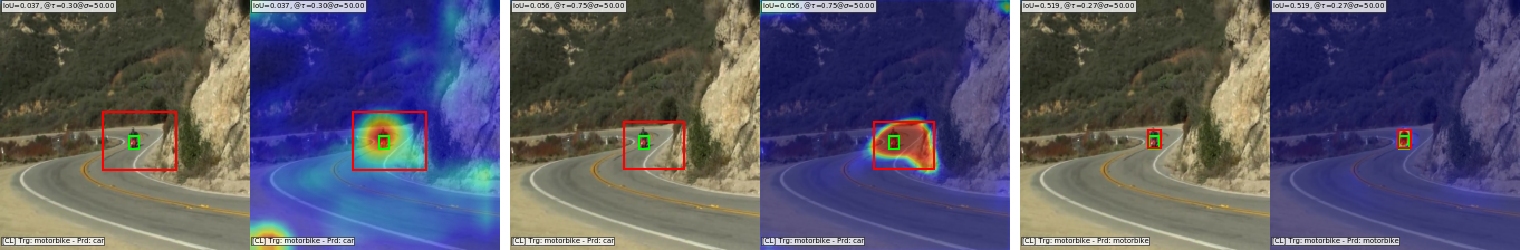} 
  \end{subfigure} 
 \\
\begin{subfigure}[b]{0.35\textwidth} 
 \centering 
 \includegraphics[width=\textwidth]{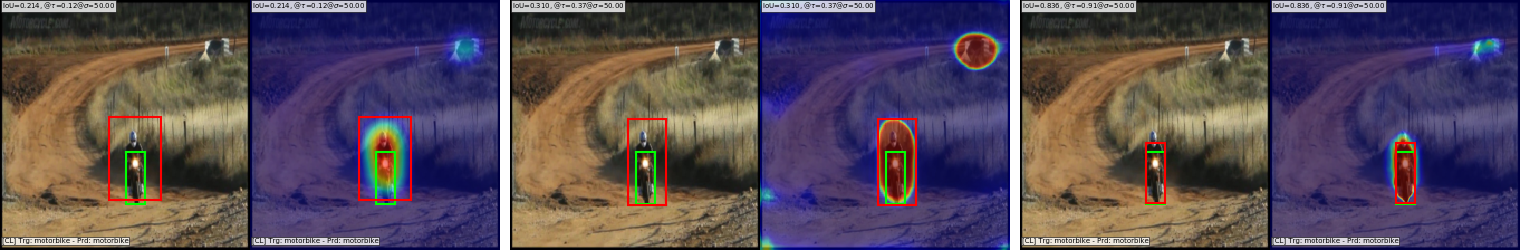} 
  \end{subfigure} 
 \\
\begin{subfigure}[b]{0.35\textwidth} 
 \centering 
 \includegraphics[width=\textwidth]{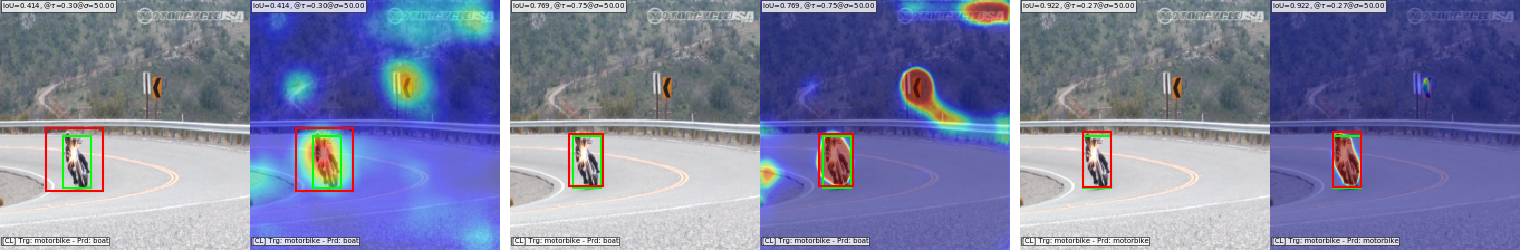} 
  \end{subfigure}
        \caption{Additional localization examples of test sets frames (\ytovone, \ytovtwodtwo). 
        \emph{Bounding boxes}: ground truth (green), prediction (red). The second column of each method is the predicted CAM over image.}
        \label{fig:visu-pred-supp-mat-1}
\end{figure}

\begin{figure}
     \centering
     \begin{subfigure}[b]{0.35\textwidth}
         \centering
         \includegraphics[width=\textwidth]{tag}
     \end{subfigure}
     \\
     \begin{subfigure}[b]{0.35\textwidth} 
 \centering 
 \includegraphics[width=\textwidth]{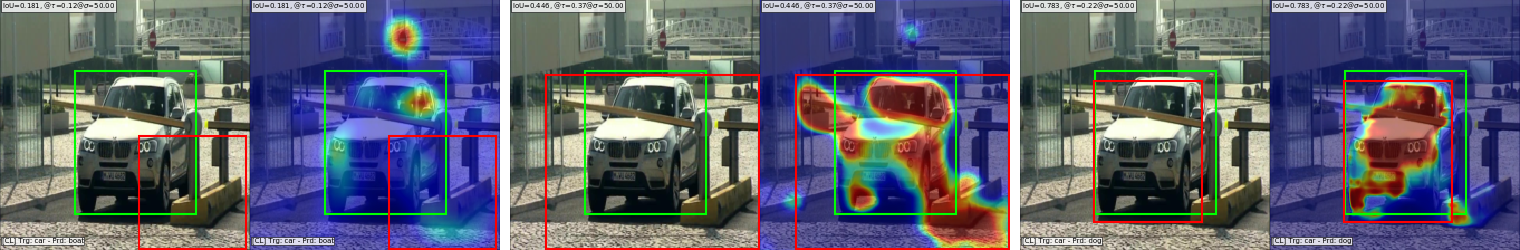} 
  \end{subfigure} 
 \\
\begin{subfigure}[b]{0.35\textwidth} 
 \centering 
 \includegraphics[width=\textwidth]{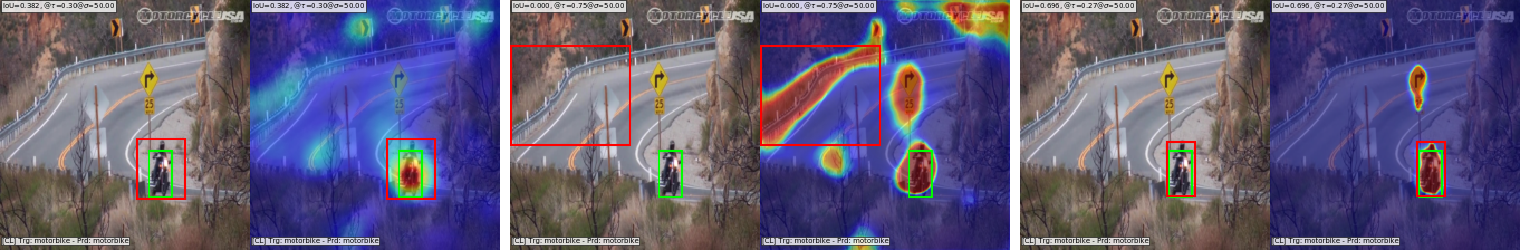} 
  \end{subfigure} 
 \\
\begin{subfigure}[b]{0.35\textwidth} 
 \centering 
 \includegraphics[width=\textwidth]{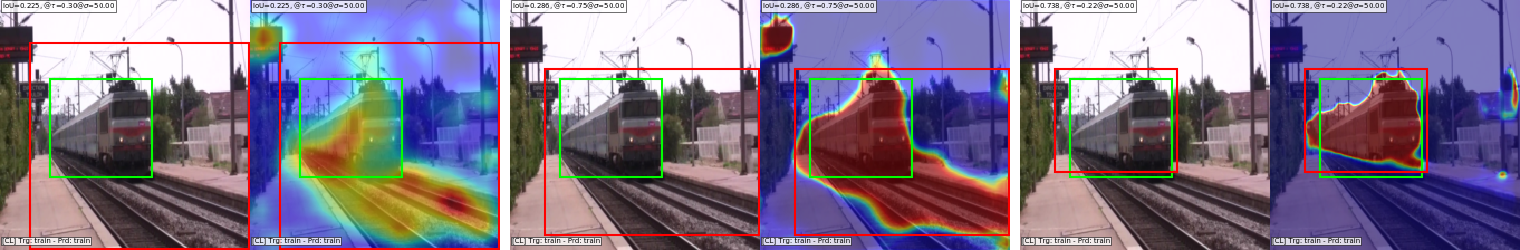} 
  \end{subfigure} 
 \\
\begin{subfigure}[b]{0.35\textwidth} 
 \centering 
 \includegraphics[width=\textwidth]{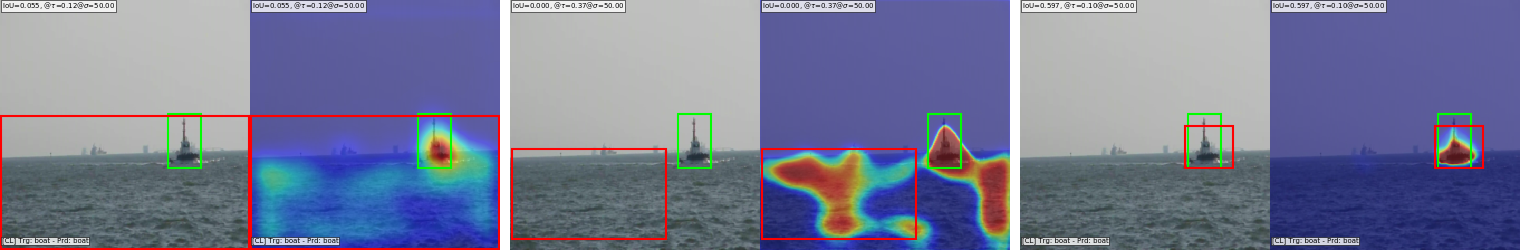} 
  \end{subfigure} 
 \\
\begin{subfigure}[b]{0.35\textwidth} 
 \centering 
 \includegraphics[width=\textwidth]{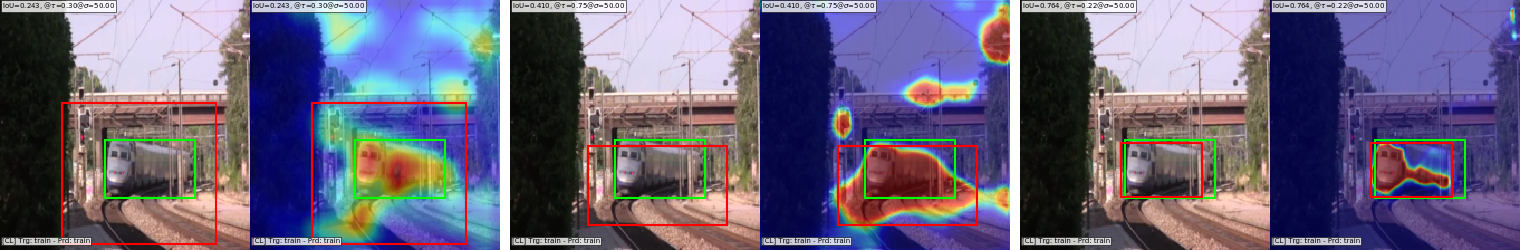} 
  \end{subfigure} 
 \\
\begin{subfigure}[b]{0.35\textwidth} 
 \centering 
 \includegraphics[width=\textwidth]{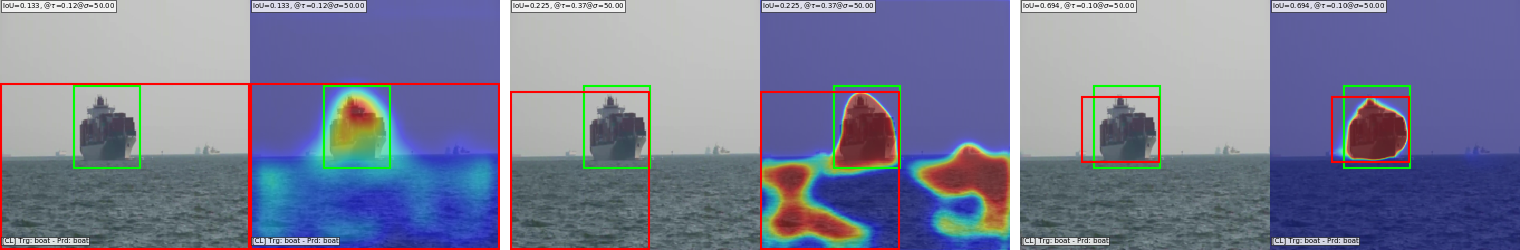} 
  \end{subfigure} 
 \\
\begin{subfigure}[b]{0.35\textwidth} 
 \centering 
 \includegraphics[width=\textwidth]{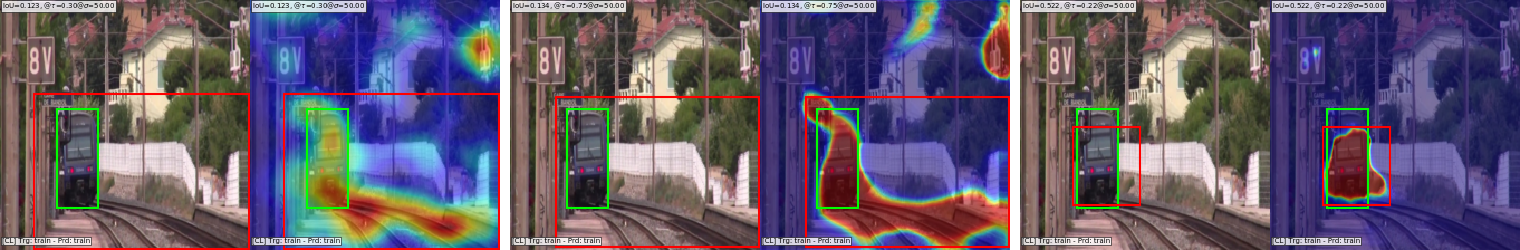} 
  \end{subfigure} 
 \\
\begin{subfigure}[b]{0.35\textwidth} 
 \centering 
 \includegraphics[width=\textwidth]{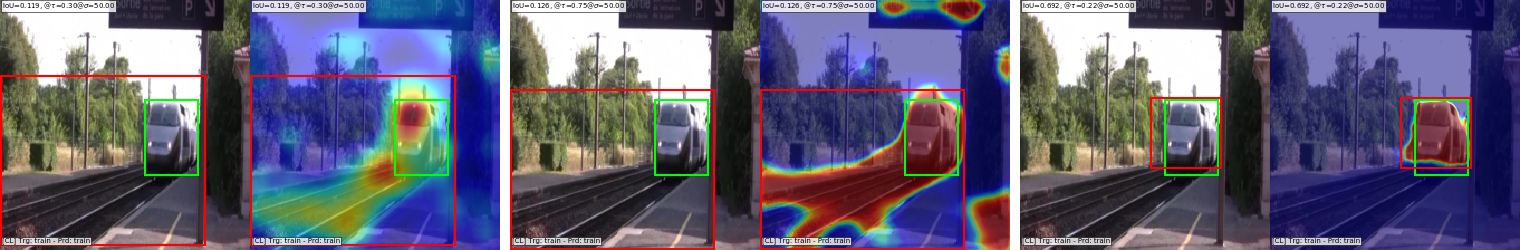} 
  \end{subfigure} 
 \\
\begin{subfigure}[b]{0.35\textwidth} 
 \centering 
 \includegraphics[width=\textwidth]{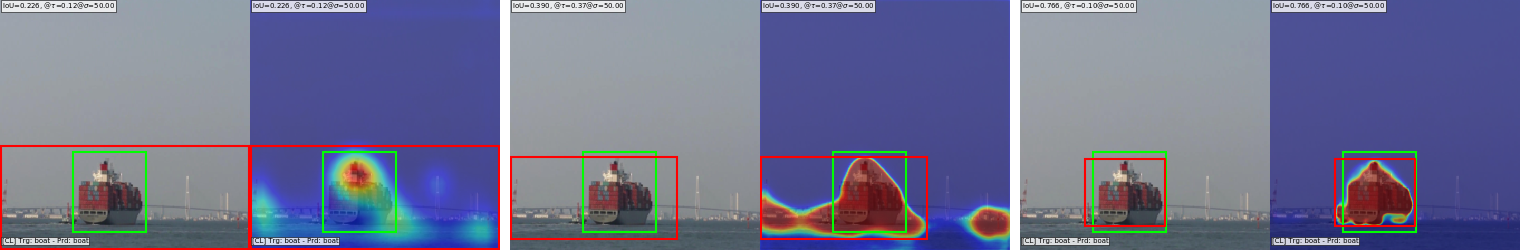} 
  \end{subfigure} 
 \\
\begin{subfigure}[b]{0.35\textwidth} 
 \centering 
 \includegraphics[width=\textwidth]{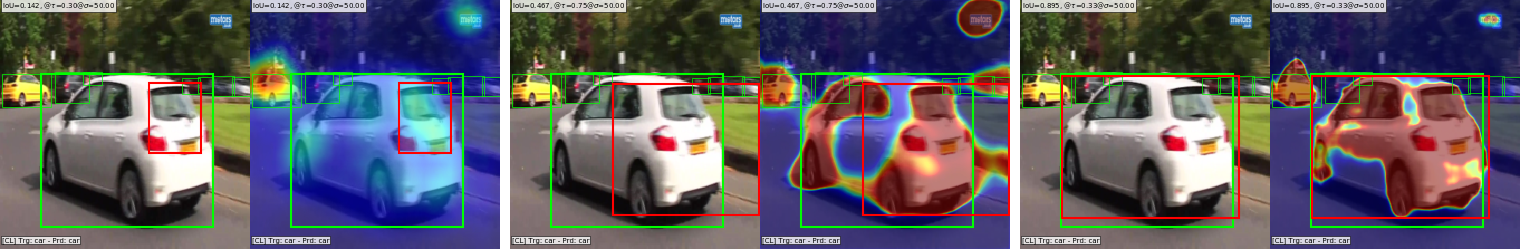} 
  \end{subfigure} 
 \\
\begin{subfigure}[b]{0.35\textwidth} 
 \centering 
 \includegraphics[width=\textwidth]{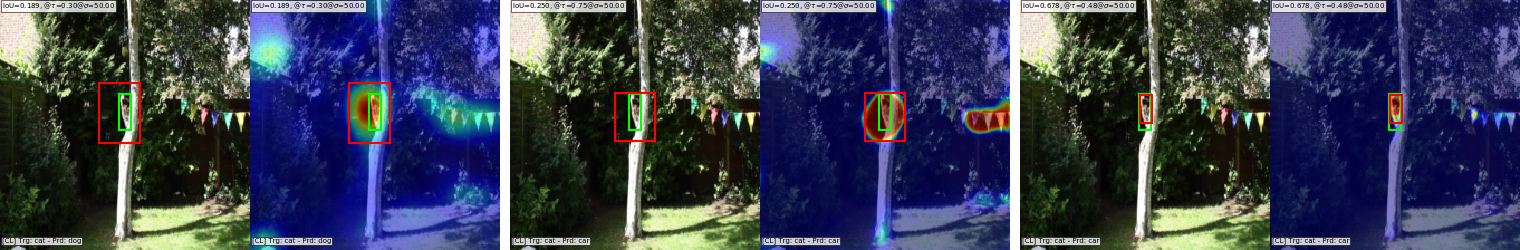} 
  \end{subfigure} 
 \\
\begin{subfigure}[b]{0.35\textwidth} 
 \centering 
 \includegraphics[width=\textwidth]{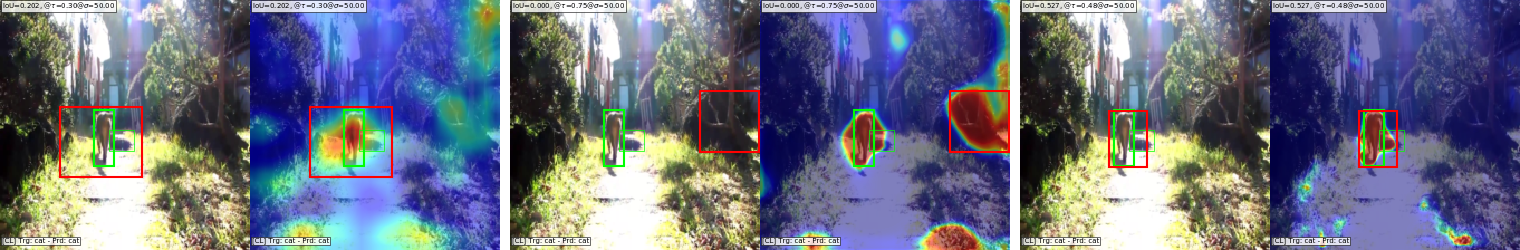} 
  \end{subfigure} 
 \\
\begin{subfigure}[b]{0.35\textwidth} 
 \centering 
 \includegraphics[width=\textwidth]{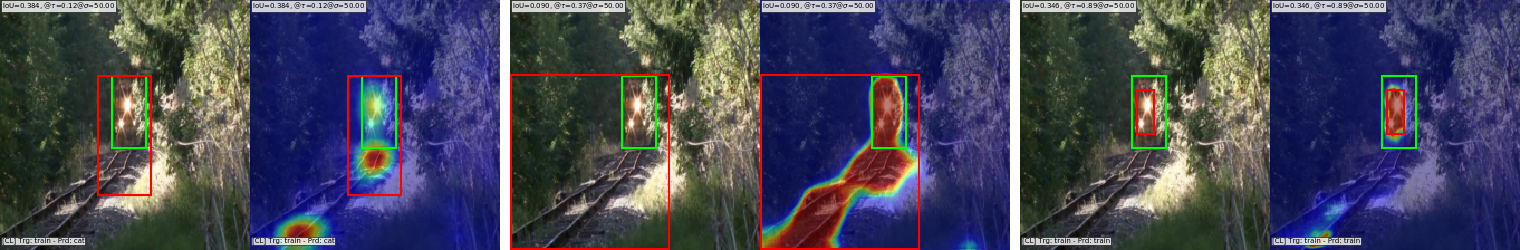} 
  \end{subfigure} 
 \\
\begin{subfigure}[b]{0.35\textwidth} 
 \centering 
 \includegraphics[width=\textwidth]{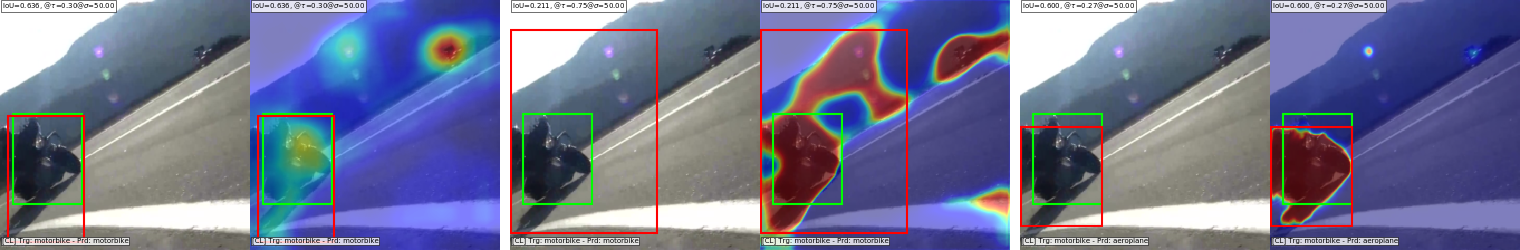} 
  \end{subfigure} 
 \\
\begin{subfigure}[b]{0.35\textwidth} 
 \centering 
 \includegraphics[width=\textwidth]{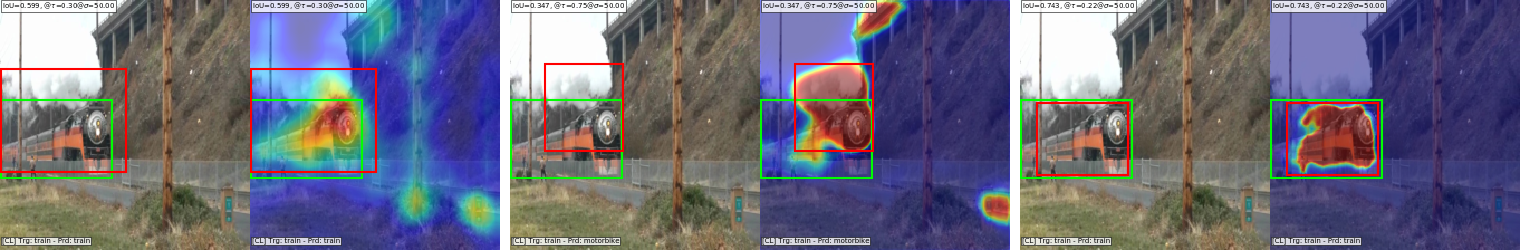} 
  \end{subfigure} 
 \\
\begin{subfigure}[b]{0.35\textwidth} 
 \centering 
 \includegraphics[width=\textwidth]{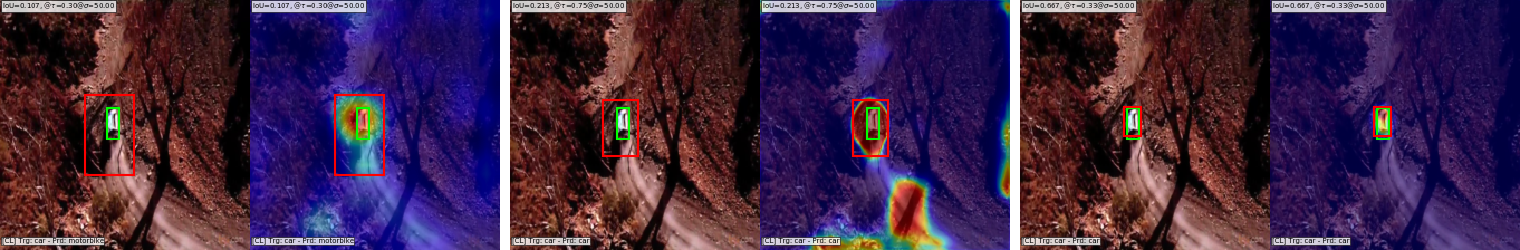} 
  \end{subfigure} 
 \\
\begin{subfigure}[b]{0.35\textwidth} 
 \centering 
 \includegraphics[width=\textwidth]{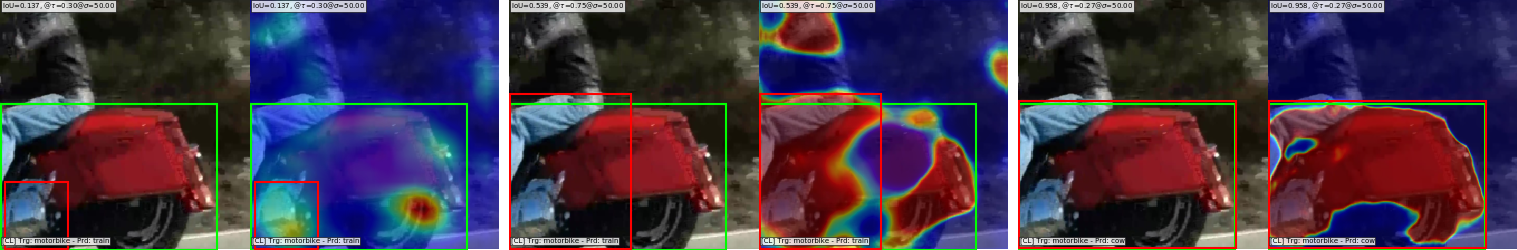} 
  \end{subfigure} 
 \\
\begin{subfigure}[b]{0.35\textwidth} 
 \centering 
 \includegraphics[width=\textwidth]{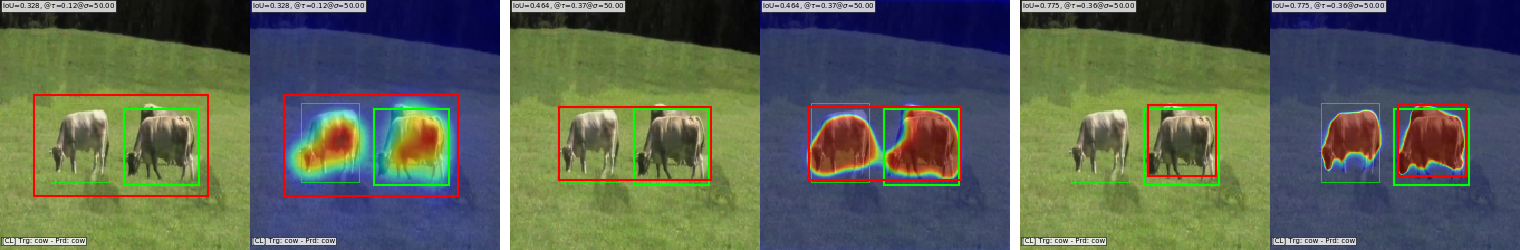} 
  \end{subfigure} 
 \\
\begin{subfigure}[b]{0.35\textwidth} 
 \centering 
 \includegraphics[width=\textwidth]{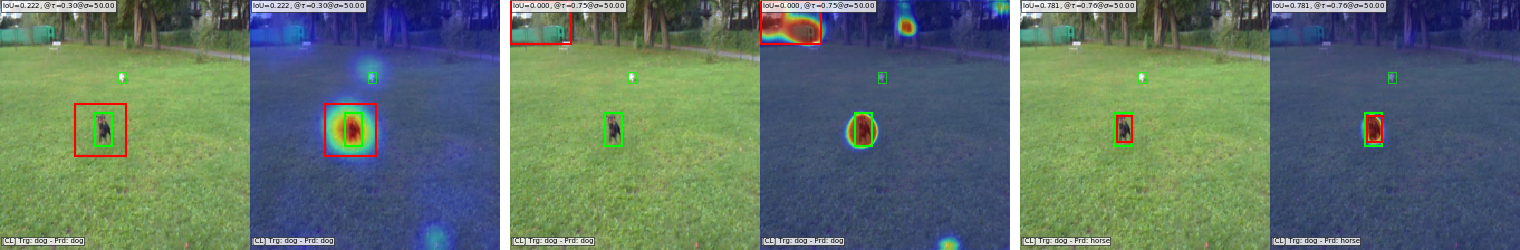} 
  \end{subfigure} 
 \\
\begin{subfigure}[b]{0.35\textwidth} 
 \centering 
 \includegraphics[width=\textwidth]{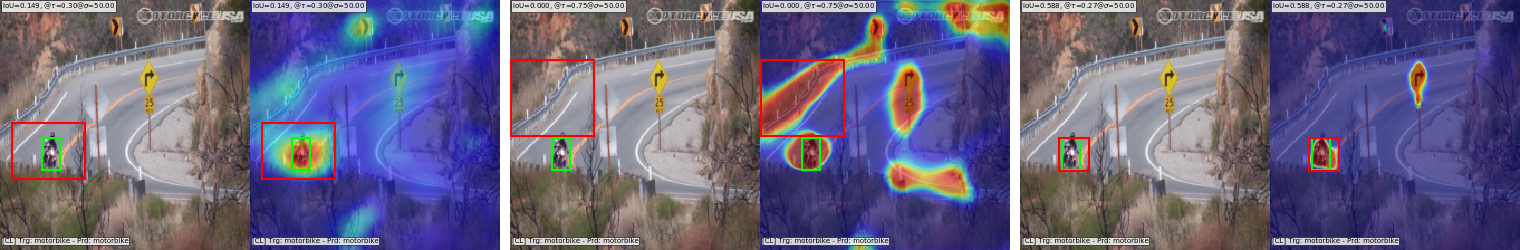} 
  \end{subfigure} 
        \caption{Additional localization examples of test sets frames (\ytovone, \ytovtwodtwo). 
        \emph{Bounding boxes}: ground truth (green), prediction (red). The second column of each method is the predicted CAM over image.}
        \label{fig:visu-pred-supp-mat-2}
\end{figure}

Fig.\ref{fig:visu-pred-supp-mat-1}, \ref{fig:visu-pred-supp-mat-2} displays more visual results. Images show that our method yielded sharp and less noisy CAMs. However, multi-instances and complex scenes remain a challenge. 

In a separate file, we provide demonstrative videos which show the localization of objects. They highlight additional challenges not visible in still images, in particular:  

\noindent - \textbf{Empty frames}. Some frames are without objects, yet still show class activations. This issue is the result of the weak global annotation of videos, and the assumption that all frames inherit that same label. It highlights the importance of detecting such frames and treating them accordingly, starting with classifier training. Discarding these frames will help the classifier better discern objects and reduce noisy CAMs. Without any additional supervision, WSVOL is still a difficult task. One possible solution is to leverage the per-class probabilities of the classifier to assess whether an object is present or not in a frame. Such likelihood can be exploited to discard frames or weight loss terms over them.

\noindent - \textbf{Temporal consistency}. Although our method brought a quantitative and qualitative improvement, there are some failure cases. In some instances, we observed inconsistencies between consecutive frames. This error is characterized by a large shift in localization where the bounding box (and the CAM activation), moves drastically. This often happens when objects become partially or fully occluded, the appearance of other instances, the scene becomes complex, or zooming in the image makes the object too large. Such setups lead to a \emph{sudden shift} of the CAM's focus, leading to an abrupt change of localization which is undesirable. This drawback is mainly inherited from the inference procedure that is achieved at a single frame and without accounting for spatiotemporal information allowing such errors. Future works may consider improving the inference procedure by leveraging spatiotemporal information at the expense of inference time.

\newpage
\clearpage

\FloatBarrier

\bibliographystyle{apalike}
\bibliography{main}

\end{document}